\documentclass[article, 3p,times,review]{elsarticle}
\usepackage{lineno,hyperref}
\usepackage[english]{babel}
\usepackage{tabularx}
\usepackage{longtable}
\usepackage{graphicx}
\usepackage[export]{adjustbox}
\usepackage{amsmath}
\usepackage{xtab}
\usepackage{multirow}
\usepackage{array}
\newcolumntype{P}[1]{>{\centering\arraybackslash}p{#1}}
\usepackage[flushleft]{threeparttable}
\usepackage{amssymb}
\usepackage{pifont}
\usepackage{mathtools}
\usepackage{xcolor}
\usepackage{enumitem}
\usepackage{amsmath}
\usepackage{kbordermatrix}
\usepackage{blkarray, bigstrut}

\usepackage{lineno,hyperref}
\biboptions{authoryear}

\begin{document}

\begin{frontmatter}

\title{Location reference identification from tweets during emergencies: A deep learning approach\tnoteref{}}


\author[First]{Abhinav Kumar\corref{cor1}}
\ead{abhinavanand05@gmail.com}

\author[First]{Jyoti Prakash Singh}
\ead{jps@nitp.ac.in}


\cortext[cor1]{Corresponding author}

\address[First]{Department of Computer Science and Engineering \\ National Institute of Technology Patna, India }
\begin{abstract}
Twitter is recently being used during crises to communicate with officials and provide rescue and relief operation in real time. The geographical location information of the event, as well as users, are vitally important in such scenarios. The identification of geographic location is one of the challenging tasks as the location information fields, such as user location and place name of tweets are not reliable. The extraction of location information from tweet text is difficult as it contains a lot of non-standard English, grammatical errors, spelling mistakes, non-standard abbreviations, and so on. This research aims to extract location words used in the tweet using a Convolutional Neural Network (CNN) based model. We achieved the exact matching score of 0.929, Hamming loss of 0.002, and $F_1$-score of 0.96 for the tweets related to the earthquake. Our model was able to extract even three- to four-word long location references which is also evident from the exact matching score of over 92\%. The findings of this paper can help in early event localization, emergency situations, real-time road traffic management, localized advertisement, and in various location-based services.

\end{abstract}

\begin{keyword}
Location references, Tweets, Geo-locations, Named entity recognition, Gazetteer, Convolutional Neural Network
\end{keyword}

\end{frontmatter}


\section{Introduction}
Tweets are very responsive to real-world events, and are sometimes even more immediate than traditional news channels. Therefore, it is possible to keep track of the latest information by following tweets. Several examples were seen when the news was first reported on Twitter, such as an airplane crash over the Hudson River in New York in the year 2009 \citep{sakaki2013tweet}, the death of former British Prime Minister Margaret Thatcher in April 2013\footnote{http://www.guardian.co.uk/technology/2013/apr/23/twitter-first-source-investment-news \label{foot}}, and the explosions at the Boston Marathon 2013\textsuperscript{\ref{foot}}. In recent years, Twitter has been used extensively in the course of natural and human-made disasters such as earthquakes, floods, fire, terrorist attacks, civil unrest, and so on \citep{alexander2014social, landwehr2016using, laylavi2017event, laylavi2016multi, luna2018social, mejri2017crisis, mendoza2010twitter, sakaki2013tweet, singh2017event, yuan2018feasibility}. The government and non-government agencies use Twitter in case of crisis so that different rescue operations can leap into action, disseminate information to the wider audience, and recognize floor reality \citep{imran2014coordinating, imran2015processing, landwehr2016using, laylavi2017event, laylavi2016multi, rossi2018early, sakaki2013tweet, zhou2017emergency}. In an American Red Cross survey, a question was asked to individuals that ``whom they contacted in an emergency?" Twenty-eight percent of Americans turned to Twitter for help if they were unable to reach the emergency contact number (911)\footnote{http://www.ehstoday.com/fire\_emergencyresponse/communications/red-cross-social-media-help-disaster-0232}. Twitter is also used in real time road traffic monitoring {\citep{gu2016twitter, hoang2016crowdsensing}, event localization \citep{giridhar2015quality, panteras2015triangulating}, and in various location-based services \citep{ikawa2012location, xu2015mobifeed}. The estimation and detection of location information of events and users from tweets are a major concern in relation to the above-mentioned tasks. \\

Twitter provides three location information fields for sharing a user's location: (1) User location; (2) Place name; and (3) Geo-coordinate. The user location field has 140 character spaces (previously it was limited to 30 characters) in which the user can write his/her home location information while creating their profile. This field is optional to the user and the user can write any arbitrary words or leave it blank. In many instances, they write meaningless words that might not refer to any location name. \cite{hecht2011tweets} analyzed that 34\% of users do not reveal their ``user location" information. \cite{cheng2010you} found that only 26\% of users use city level or below city level location names in their user location field. However, this field can not be treated as the current location of the user as it is entered at the time of creating their profile and most of the time not updated by the users regularly. The second field is for the ``place name," which can be attached to a tweet when it is posted. The place name is represented by a location name with an array of the latitude-longitude pair in the form of the location's boundary coordinates. These place names are predefined on the Twitter database, but it does not provide granular location information. \cite{kumar2017authenticity} found that only 47.33\% of tweets contain place names. However, 12\% of those place names are incorrect in terms of their spatiotemporal information. The third field provided by Twitter is for the ``geo-coordinates" (geographical footprints of latitude and longitude) that can be attached at the time of posting a tweet using a GPS- (Global Positioning System) enabled device. Most of the researchers \citep{huang2014tweets, nakaji2012visualization, yuan2013and} have considered geo-coordinates as the most explicit and precise information, i.e., tweets associated with latitude-longitude information. However, tweets with geo-coordinate information are infrequent. \cite{cheng2010you}, \cite{morstatter2013sample}, and \cite{kumar2017authenticity} determined that only 0.42\%, 3.17\%, and 7.90\% of tweets respectively are geo-tagged. \cite{kumar2017authenticity} further reported
that although geo-coordinates are the most precise location information, they are not always authentic in terms of their spatiotemporal information if the tweet is posted from third-party applications such as Instagram\footnote{https://www.instagram.com/} etc. Hence, all three location information fields, available in tweets and user accounts, have their own limitations and cannot be completely relied on.\\

Along with the location fields mentioned above, people also make location references in their tweet texts when asking for help or reporting the event of a disaster. It is found that people from a disaster-related area tend to use their location information in their tweet text \citep{vieweg2010microblogging}. The available location information in tweet texts is vitally important as it represents the location information of any event or user during emergencies. Hence, the location information mentioned in the tweet text may be considered as the most authentic source of geographic evidence in an emergency. The tweet text is a free-text field limited to 280 characters (previously it was 140 characters). Location information from these tweet texts can be extracted using either the gazetteer-based approach \citep{itoh2016spatio, li2014fine, malmasi2015location, middleton2014real, sankaranarayanan2009twitterstand, zhang2014geocoding} or the Named Entity Recognition (NER) based approach \citep{gelernter2011geo, giridhar2015quality, unankard2015emerging}. Gazetteer is a corpus of location names (e.g., GeoNames\footnote{http://geonames.org}). In the gazetteer-based approach, the words of tweets are looked up in the gazetteer to find the location names. However, there are some inherent problems with this approach: (i) the unavailability of gazetteers for all the regions; and (ii) a location name mentioned in the text may have some other non-geographic meaning in the context of a text e.g., the word ``Reading" may refer to a location name in England or it may also be used in another context. The other problem with this approach is the geo-ambiguity (distinct locations have the same name, e.g., Paris has 140 possibilities). The second approach is Named Entity Recognition (NER). The NER technique generally tokenizes each word of the tweet using language-specific part-of-speech tagging, then it detects the group of words that probably refer to named entities. This approach works well for well-written English sentences, but it does not work well for tweet texts as they have several grammatical mistakes, nonstandard abbreviation, and spelling mistakes \citep{ajao2015survey, gelernter2011geo, ozdikis2017survey, zheng2018survey}. \cite{temnikova2015case} did an extensive analysis on the readability of tweets during the crisis and suggested several recommendations for writing understandable tweets. In many cases, a number of English language rules are violated e.g., the first letter of the proper nouns are not usually written in capital letters. Also, the grammar is not correct in many scenarios e.g., missing prepositions. Further, most users do not use the correct spelling in their tweets. They often write words in short by removing the vowels from words. To resolve the aforementioned problems and find the location references, several efforts have been made by researchers, such as \cite{lingad2013location}, who re-trained the Named Entity Recognition tool for the Twitter environment, \citep{li2012twiner, liu2011recognizing, ritter2011named}, they re-built their own Named Entity Recognition framework. Some other works also combined the gazetteer and NER approaches to find named entities from tweets \citep{gelernter2013algorithm, middleton2018location}.\\

In most of the earlier work, NER-based approaches used POS tagging and extracted all named entities such as person name, product, group, corporation, location, etc. In the current work, we are concentrating on the extraction of location words ignoring other named entities. For this, instead of using POS tagging, we train a Convolutional Neural Network- (CNN) based system to extract location names present in the tweet. We represent the tweet text as normal sentences and highlight the words containing location information. We assume that there is already a system that filters tweets based on their relatedness to a particular event. Several works have been reported regarding this \citep{chowdhury2013tweet4act, imran2014coordinating, nguyen2017robust, olteanu2014crisislex, singh2017event}. Once the tweets are found to be related to the event, our model finds the location referring words in that tweet. We present this problem under the supervised learning paradigm. A dataset of tweets and their corresponding location words are made to train a system. Since the input is a sentence (tweet text), we had several options, such as LDA \citep{blei2003latent}, PLSA \citep{hofmann1999probabilistic}, and word embedding \citep{pennington2014glove} to represent the sentence. LDA and PLSA are generative statistical models that can represent a document as a mixture of a small number of topics. They are widely used for grouping tweets related to a specific event. Our target is to preserve the sentence structure so that the corresponding word number can be marked as a location word or not. This is why we prefer word embedding over other techniques, such as LDA or PLSA.\\

As a supervised learning model, there are several options starting from simple machine-learning models, such as SVM, Naive Bayes, Random Forest to deep-learning models, such as the Recurrent Neural Network (RNN), the Convolutional Neural Network (CNN). The machine-learning model requires some features to learn to associate input with output. Therefore, the performance of these systems heavily depends on the feature engineering. This is why we choose deep learning models. RNNs are good for sequential or long-text data. Tweets have short sentences, which favor the use of CNN over RNN. The intuition behind using CNN is that the convolutional layer can automatically learn the better representation of input data and then dense layers can utilize these input representations to identify location references. Our objective for the current work is formulated as: (i) Find whether a tweet contains a location name; and (ii) If there are location names present in a tweet, then highlight those words. \\

The remainder of the paper is organized as follows: in section \ref{RW}, we briefly present the related literature. Our proposed framework is presented in section \ref{Meth}. The finding of the proposed system is presented in section \ref{Res}. Section \ref{Dis} contains discussion about the results and implications of the current research. We conclude the paper in Section \ref{Con}.

\section{Related work}
\label{RW}
Recently, a number of works have been reported for better utilization of social media for emergency purposes \citep{ajao2015survey, alexander2014social, imran2015processing, ozdikis2017survey, shibuya2017mining, zheng2018survey}. \cite{olteanu2015expect} investigated several natural hazards and human-induced disasters in a systematic way to better understand the effective use of social media for information gathering processes during emergencies. Most of the existing works focus on event detection and location estimation of events or users during emergencies. In event detection, some researchers tried to detect an event as soon as possible, whereas some researchers tried to classify event-related tweets into predefined classes to conduct further analysis. In the location estimation, researchers tried to find the location of the events or users from social media. We are dividing this section into two subsections to better organize the existing works: (i) Event detection; and (ii) Location estimation. 

\subsection{Event detection}
\cite{imran2013extracting} proposed a system that used machine-learning methods to detect informative messages during a crisis. After detecting informative messages, their system automatically extracted nuggets of information from them. \cite{olteanu2014crisislex} proposed a methodology for building an effective lexicon for crisis events. Their approach could improve the recall in the sampling of Twitter communication, which could greatly help in situation awareness during a crisis. \cite{imran2014aidr} proposed an AIDR (Artificial Intelligence for Disaster Response) platform for automatic classification of disaster-related messages into user-defined classes. \cite{chowdhury2013tweet4act} used content-based features, such as n-gram and the tense of the message to automatically classify messages into pre-incident, during-incident, and post-incident classes. \cite{perol2018convolutional} used a convolutional neural network for earthquake detection and location estimation from seismograms. \cite{nguyen2017real} proposed a convolutional neural network based model for classifying earthquake-related tweets into informative or non-informative classes. Their system could detect earthquake events earlier than the announcement from the official government website. \cite{yang2017harvey} used several classifiers to identify tweets related to flood victims and volunteers. They then proposed first come first served, static priority, and hybrid rescue-scheduling algorithms to provide help for victims as soon as possible. The extensive survey on event-detection techniques for Twitter can be seen in \citep{atefeh2015survey}. 

\subsection{Location estimation}
\cite{jurgens2015geolocation} presented a comprehensive analysis of network-based approaches for predicting the geo-location of users. \cite{do2017multiview} proposed a deep multiview learning model that combines the textual, network, and metadata features for predicting the geo-location of a tweet. \cite{qian2017probabilistic} proposed a probabilistic model, integrating the content and network features learned from social media, to predict the location of a user. \cite{lourentzou2017text} utilized neural network architecture to predict the geo-location of users. They found that the choice of appropriate network architecture and hyper-parameter selection can give better accuracy in predicting geo-location. A group of researchers \citep{do2017multiview, laylavi2016multi, qian2017probabilistic} used tweet text with other metadata such as ``user location", ``geo-coordinates", and ``place name" to estimate location information. Another group \citep{gelernter2013algorithm, lingad2013location, malmasi2015location} used only tweet texts to find location information. The tweet text is used because (i) geo-tagged tweets are infrequent, (ii) the user location field is not treated as the current location of the users as this field is mostly outdated. Gazetteer and Named entity recognition-based approaches are common techniques for finding location references from tweet texts. We categorize this section in three subsections: (i) the studies using gazetteers for finding location references; (ii) the studies using Named Entity Recognition for finding location references; and (iii) the studies to resolve the issue of noisy text by developing new methods.

\subsubsection{Gazetteer Based Approach}
\cite{middleton2014real} used gazetteers, street maps, and volunteered geographic information to develop real-time crisis-mapping by geoparsing the tweet text. They used 2,000 human labeled tweets to evaluate their results and found high precision for street-level location names. They stated that a high precision (0.90 or, above) can be found in location finding from the real-time tweets by preloading location information for the area that is at risk of any disaster. \cite{sankaranarayanan2009twitterstand} clustered several different news topics and applied a Part-of-Speech (POS) tagger and a Named Entity Recognizer to find location names from the tweet texts. However, they concluded that the Named Entity Recognizer fails to give good results in the case of tweets because it is difficult for the system to work efficiently over noisy text. Further, they applied TF-IDF to extract key phrases from tweets and used GeoNames\footnote{http://geonames.org} gazetteer to find the location names. They used the extracted location names with other metadata to assign the location to clusters. \cite{itoh2016spatio} built their own gazetteer from geo-located posts submitted by the users from location-based services such as Foursquare. They enriched the gazetteer by adding named entities referring only to specific locations. They also used the parts-of-speech tagger to tag proper nouns from the text and added them to the list of the specific locations. They obtained 38,504 entries in their gazetteer to map a spatiotemporal visualization of the sports games and earthquake events. \cite{malmasi2015location} proposed an unsupervised approach based on the Noun Phrase extraction and n-gram-based matching using the GeoNames gazetteer. They claimed that their system is better for the noisy microblog text. They used 2,000 manually annotated tweets to train the system and tested it with 1,000 tweets. They achieved an $F_1$-score of 0.792. \cite{li2014fine} proposed a framework \textit{named PETAR}, which included two components, one is the Point of Interest (POI) inventory and the second is a time-aware Point of Interest (POI) tagger. The POI inventory is built using the Foursquare check-ins, which consist of formal names of POI as well as informal abbreviations. The POI tagger is based on the Conditional Random Field model. They performed their analysis on 4,000 manually labeled tweets and achieved an $F_1$-score of 0.87. \cite{zhang2014geocoding} used supervised machine-learning algorithms and utilized the gazetteer to build a model. They evaluated their model on 956 manually labeled tweets to find the location references mentioned there. \cite{al2017location} proposed a system that extracts the location names from the text using n-gram statistics and location name gazetteers. Their location name extraction tool used augmented and filtered region-specific gazetteers to detect boundaries of multi-word location names.
  
\subsubsection{Named Entity Recognition}
\cite{unankard2015emerging} used clustering on tweets and then used a Standford Named Entity Recognizer \citep{ritter2011named} to extract location names from the tweet text. They found correlations between user location and event location to localize events such as the Indonesian earthquake and the Queensland election 2012. Finally, they found the most frequent location names present in the cluster and took it as the location of the event. \cite{giridhar2015quality} used road traffic twitter data from three major cities in California and clustered the tweets mentioning an event in a specific group. They tokenized each tweet and tagged each word using a Part-of-Speech (POS) tagger to find location names. Besides POS tagging, they observed that the location names were preceded by prepositions such as \textit{in}, \textit{around}, \textit{between}, and \textit{after}. In addition, they applied this grammar-based rule to find location names. After extracting location names, they obtained the geo-coordinates of each of the extracted location names using Google maps API\footnote{https://cloud.google.com/maps-platform/} and then averaged them to find the centroid of the events. \cite{gelernter2011geo} conducted their study on the Stanford Named Entity recognizer\footnote{https://nlp.stanford.edu/software/CRF-NER.html} to know the effectiveness of it in finding location information from the tweets. They found the Stanford Named Entity could find the location names that are proper nouns, but fails to recognize local street names, buildings, nonstandard place abbreviations, misspellings and location names not starting with a capital letter. They commented that the result should improve if several named entity recognition algorithms are configured to work together. \cite{sikdar2016feature} proposed a named entity system for extracting the named entity from the tweets and then classifying those names' entities in the ten different classes. For the named entity recognition they used several lexica, character, and context-based features of the tweets. Their system achieved an $F_1$-score of 0.63 for named entity recognition and 0.40 for named entity classification. 

\subsubsection{Efforts for Twitter Named Entity Recognition}
Some researchers tried to train existing NER systems with related social media text to better learn the named entities mentioned in them. \cite{lingad2013location} used several named entity recognizers namely, Stanford NER\footnote{http://nlp.stanford.edu/software/CRF-NER.shtml}, OpenNLP\footnote{http://opennlp.apache.org}, Yahoo! PlaceMaker\footnote{http://developer.yahoo.com/boss/geo/}, and TwitterNLP \citep{ritter2011named} to find the location names from the disaster-related tweets. They retrained Stanford NER and Open NLP using the disaster-related tweets. They achieved an $F_1$-score of 0.902 for the re-trained Standford NER and $F_1$-score of 0.833 for Open NLP. \cite{li2012twiner} developed a novel two-step unsupervised Name Entity Recognition system named \textit{TwiNER} using Wikipedia and Web N-Gram corpus. Their \textit{TwiNER} named entity recognizer achieved comparable performance with other conventional NER systems for the real-life targeted tweet stream. They achieved $F_1$-scores of 0.772 and 0.419 for the two different ground-truth labeled datasets. \cite{ritter2011named} experimented with the conventional NER tools and found that the accuracy dropped from 0.97 to 0.80 when it was applied to news and tweet corpus respectively. They addressed this problem by rebuilding the NLP pipeline starting with POS tagging, through chunking, to named-entity recognition. \cite{gelernter2013algorithm} used open-source Named Entity Recognition software and machine-learning techniques to identify location references, such as streets, addresses, buildings, location names, place acronyms, and abbreviations. To identify streets, buildings, and location names they used lexico-semantic pattern recognition, Named Entity Recognizer, and gazetteer, respectively. They found an $F_1$-score of 0.85  for streets, 0.86 for buildings, 0.96 for location names, and 0.88 for abbreviated place names. Overall, they found an $F_1$-score of 0.90 in identifying location references. \cite{middleton2018location} proposed two different location extraction techniques: (i) entity matching by utilizing the OpenStreetMap database; and (ii) language model that makes use of numerous gazetteers and a large social media tag dataset. They also experimented with three different models that used third-party applications, such as GeoNames, Google Geocoder API, etc. They found that the OpenStreetMap database performed better among all five approaches with $F_1$-scores between 0.90 and 0.97 for the English and Italian tweets and an $F_1$-score of 0.66 for Turkish tweets. \cite{liu2013named, liu2011recognizing} proposed a named entity recognition framework combining the three components, which are tweet normalization, K-Nearest Neighbors (KNN) with a linear Conditional Random Fields, and a semi-supervised framework. They performed their analysis on 12,245 manually labeled tweets and found the overall $F_1$-score of 0.80 in finding named entities, such as the person, product, location, and organization. \cite{dutt2018savitr} proposed a system to infer location names mentioned in the text of tweets in an unsupervised fashion. They applied several preprocessing on tweets and then used a POS tagger to find proper nouns. After that, they used a gazetteer-based approach to find the location names mentioned in the tweet text with an $F_1$-score of 0.79. The deep neural models are also used for Named Entity Recognition by several researchers \citep{chiu2015named, collobert2011natural, huang2015bidirectional,  lample2016neural}. \cite{limsopatham2016bidirectional} used bidirectional Long Short-Term Memory (LSTM), which learns the orthographic features of tweets. They extracted both character-based word representation and word-vector representation corresponding to each word of the tweet and found an $F_1$-score of 65.89 in finding named entities. \cite{von2017transfer} used transfer learning and sentence level features for named entity recognition on tweets and achieved an $F_1$-score of 40.78. 

Most of the earlier works used Named Entity Recognition (NER) and gazetteer-based approaches to find the location information from the tweet text. Existing works require a predefined set of features and location-specific gazetteers as the input for extracting location information. Therefore, the performance of these systems are heavily dependent on feature engineering. We are eliminating the feature extraction and POS tagging by using the deep Convolutional Neural Network (CNN) to find the location references mentioned in the tweets.

\section{Methodology}
\label{Meth}
The proposed convolutional neural network-based model learns the continuous representation of tweets and then picks salient features from them to predict the location names present in the tweets. The proposed architecture has three parts: (i) word embedding that represent tweets in the vector form; (ii) convolutional model that learns the salient features from the tweets representation; and (iii) a fully connected layer that interprets the extracted features to predict the output. The detailed proposed architecture is presented in Figure \ref{MD}.

\subsection{Data Collection, Preprocessing, and Labeling}
We collected tweets related to earthquakes using the keywords \textit{earthquake} and \textit{\#earthquake} from Twitter streaming API \footnote{https://developer.twitter.com/en/docs}. \cite{olteanu2014crisislex} proposed a system for selecting the keywords to extract relevant tweets for social media during an emergency. The data collection was accomplished between 20th October 2017 to 15th March 2018 for several earthquakes across different parts of the world, such as Iran, Mexico, Iraq, the Philippines, New York, Algeria, the United States, and Peru, to name a few. We collected a total of 103,384 tweets related to earthquakes in JSON (JavaScript Object Notation) format. The tweets contained the tweet text along with metadata, such as the posting time of tweets, user ID, tweet ID, and so on. We randomly selected a subset of these tweets to annotate the location references mentioned in the tweet text \citep{karimi2012microtext}. We kept the tweet text only and discarded other metadata for the current work as we wanted to focus on finding location words in the text only. We pre-processed the tweets to first remove non-English tweets and then removed duplicate tweets, mentioned user names, URL links, and emoticons. The duplicate tweets were removed by finding RT (re-tweets) in the tweet text. Hashtags were replaced with the corresponding word (e.g., \#Mexico to Mexico). The text was converted to lower case. The stopwords were kept in the tweet as their occurrence may indicate the start of location words. We kept all the words of the entire tweet collection to make word representation even if they occurred only once. After pre-processing, the dataset has only the tweet text without any user identification marks. Hence, the user privacy has not been compromised in the current research.

In our dataset, tweets have a diverse granularity of location references, such as street name, building name, city, district, and even country name. We observed that several tweets have more than one piece of location information, i.e., tweets with multiple location references. Some location names need more than one word to represent it in the tweet. For example, this tweet ``\textit{I had the same experience with the earthquake in New York back in 2012. I felt my office shake but nobody knew what happened until I saw Twitter}" has two words: \textit{New} and \textit{York}, to refer the location name \textit{New York}. Three postgraduate students volunteered to annotate the location references mentioned in the tweets. They individually annotated the tweet for words related to location references. We considered only those location references on which at least two students agreed. Finally, we obtained a total of 5,107 annotated tweets with 6,690 location references; a detailed description of the dataset used in this study is listed in Table \ref{dataset}. The sample tweets with location references and their annotations are listed in Table \ref{tweet_info_up}.

\begin{table}
	\caption{Description of the tweets containing words referring to location names}
	\begin{center}
		\label{dataset}
		\begin{tabular}{ | P{3cm} | P{4cm} |P{5cm}|}
			\hline
			Number of Tweets (Total = 5,107) &	Number of words referring to location names in a tweet & Total number of words referring to location names (Total = 6,690)   \\ \hline
			
			1897 & 0 & 0\\ \hline
			
			1300  & 1 & 1300\\ \hline
			
			1016 & 2 &  2032\\ \hline
			
			499  & 3 & 1497 \\  \hline
			
			240 & 4 & 960 \\ \hline
			
			155  & $\geq$ 5 & 901 \\  \hline
		\end{tabular}
	\end{center}\
\end{table}

\begin{table}[h]
	\caption{Sample tweets containing words referring to location names}
	
	\begin{center}
		\label{tweet_info_up}
		\begin{tabular}{ | p{12cm} | p{2.8cm} |}
			
			\hline
			
			Tweets &	Location references  \\ \hline
			
			\texttt{Hey @AppleSupport my friend @carloxito lost everything in \#Mexico \#earthquake, incl his iMac. Can you help him fix? http://bit.ly/2yA8HHI} & \texttt{Mexico} \\ \hline

			\texttt{Help out! Give to 'Relief for Earthquake Victims in Kurdistan'. https://www.generosity.com/fundraisers/2269042 … \#generosity via @generosity}  & \texttt{Kurdistan} \\ \hline
			
			\texttt{Moderate earthquake, 5 mag has occurred near Maasin in Philippines - https://wp.me/p5bFdp-rQQ  \#earthquake \#quake} & \texttt{Maasin, Philippines} \\ \hline
			
			\texttt{Small earthquake felt here, Missouri, Tennessee... https://fb.me/2K4lPX0f1} & \texttt{Missouri, Tennessee} \\ \hline
		
			\texttt{There was an earthquake of seismic intensity 4 in Tokyo earlier. There is no damage.💥} & \texttt{Tokyo}  \\ \hline
			
			\texttt{Earthquake hits central Iraq, felt in Baghdad - Reuters http://fxmb.info/Q9m1rY  \#hng \#earthquake http://earthcentral.org} & \texttt{Iraq, Baghdad}  \\ \hline

			\texttt{I had the same experience with the earthquake in New York back in 2012. I felt my office shake but nobody knew what happened until I saw Twitter} & \texttt{New York} \\ \hline

		\end{tabular}
	\end{center}\
\end{table}

\begin{figure}
	\begin{center}
		\fbox{\includegraphics[scale =.85]{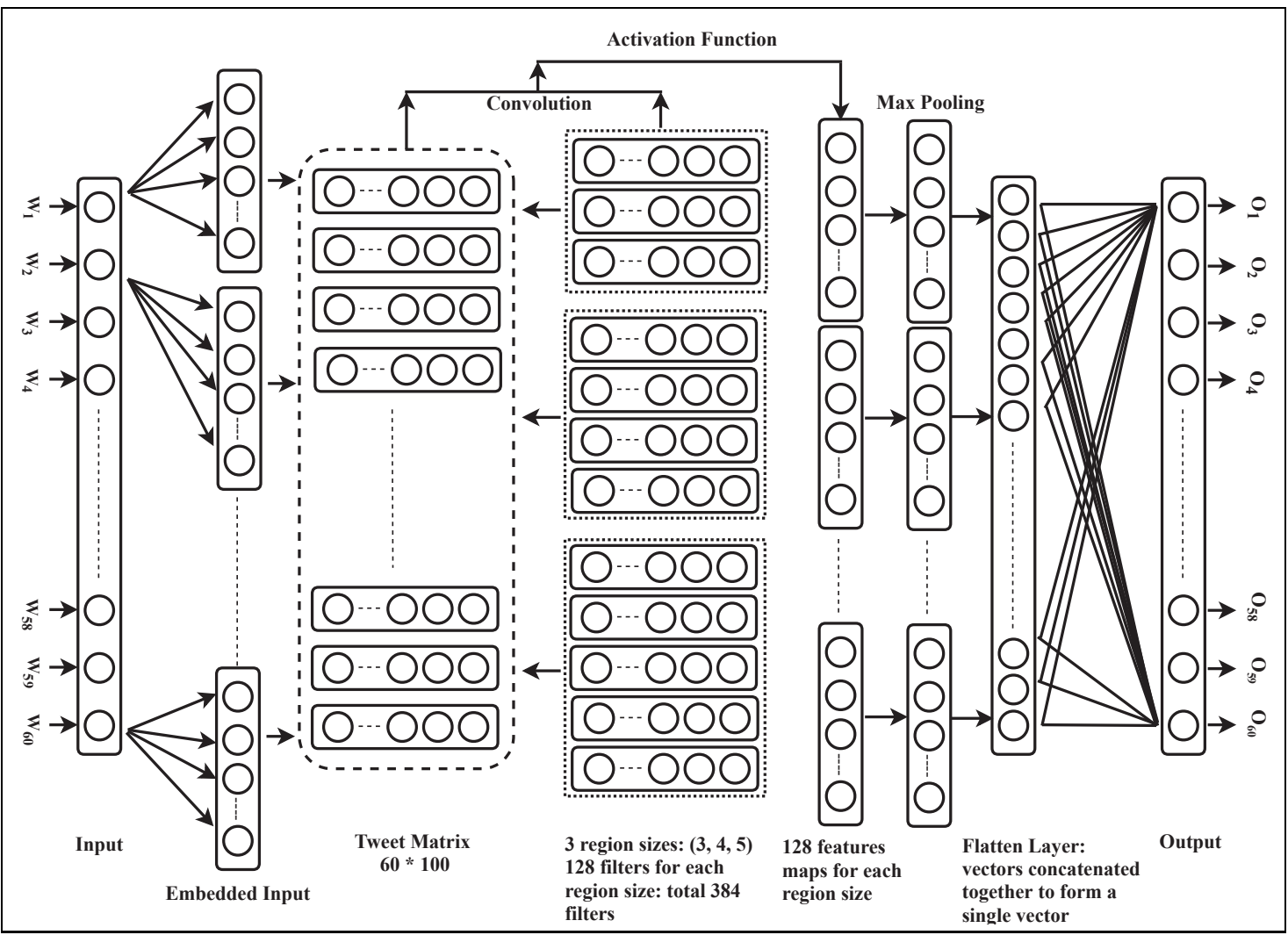}}
		\caption{Overall architecture for the Convolutional Neural Network (CNN)}
		\label{MD}
	\end{center}
\end{figure}

\subsection{Word embedding to represent tweets}
We used word embeddings of tweets as the input to the model. Word embedding represents the real-valued vector representation of the words of the text corpus in a predefined fixed dimension. The word embedding creates similar vectors for words with similar meanings. For the representation of the word vector, we created a bag-of-words $W$ from all unique words in our tweet texts. After that, for each word $W_i$, we made a look-up matrix $L_M$ to get its embedding in the $K$ dimensional vector, represented by $e(W_i) \in R^K $ , where $L_M$ represents $|W| \times K$ dimensional $(R^{|W| \times K})$ vector. Basically, two types of initialization can be done to represent the look-up matrix $L_M$. First, for the look-up matrix $L_M$, all word vectors can be randomly initialized from a uniform distribution \citep{socher2013recursive}. Second, it can be a pre-trained word vector from a large corpus of text using the embedding learning algorithm \citep{mikolov2013distributed, pennington2014glove}. In our case, we used the pre-trained word embedding GloVe (Global Vectors for word representation) \citep{pennington2014glove} as the look-up matrix for the experiment. Each word of the tweet is fed as input to the \textit{Embedding Input} through the input layer, where the weight matrix between the input and embedding layer is the pre-trained look-up matrix $L_M$. We used GloVe (\textit{`glove.twitter.27B.100d.txt'\footnote{It is freely available at https://nlp.stanford.edu/projects/glove/}}) with 100 dimension vectors embedding trained by the \textit{Google} on 27 billion words of the tweets. The use of pre-trained GloVe embedding reduces the computation overhead and normally offers better results as it is trained over the massive corpus of the texts \citep{goldberg2016primer}.\\

To represent the complete tweet in its matrix form, we concatenated the embedding of each word of the tweet. Suppose $T_i$ represents a tweet of length $m$ (padded where necessary), the complete tweet in a matrix form can be represented by equation \ref{t_matrix}. 

\begin{equation}
\label{t_matrix}
T_i = W_{1:m} = e(W_1) \oplus e(W_2)\oplus e(W_3)\oplus....\oplus e(W_m)
\end{equation} 

where, $\oplus$ represents the concatenation operator, $W_{1:m}$ represents the concatenation of word from $1$ to $m$. Padding is used to fix the length of each tweet to the same size. The complete tweet matrix $T_i$ is represented as given below: 

\[
T_i = \kbordermatrix{
	& W_1 & W_2 & W_3 & \dots & W_m \\
	 & \alpha_{11} & \alpha_{21} & \alpha_{31} & \dots & \alpha_{m1}\\
	  & \alpha_{12} & \alpha_{22} & \alpha_{32} & \dots &\alpha_{m2}\\
	   & \alpha_{13} & \alpha_{23} & \alpha_{33} & \dots &\alpha_{m3}\\
	 & \vdots & \vdots & \vdots & \vdots & \vdots \\
	  & \alpha_{1K} & \alpha_{2K} & \alpha_{3K} & \dots & \alpha_{mK}\\
}
\]

where, $[\alpha_{m1}  \hspace{0.1cm} \alpha_{m2} \hspace{0.1cm}  \alpha_{m3} \hspace{0.1cm}  ..... \hspace{0.1cm} \alpha_{mK}] $ represents the embedding of word $W_m$, tweet matrix $T_i$ having $m$ words of dimension $R^{m \times K}$. The pictorial view can be seen from Figure \ref{MD}, where each embedding input of a word is concatenated one after the other in a sequence to form a complete tweet representation. This tweet representation is then used by the Convolutional Neural Network (CNN) to learn the location references present in the tweet.

\subsection{Context dependent feature extraction}
The convolution process of the CNN model is used to extract the semantic features of the sentence \citep{kalchbrenner2014convolutional, socher2013recursive} by using n-gram information \citep{ collobert2011natural}. In the CNN model, the convolution process involves a filter $F \in R^{h \times K}$, with the size of $h$ words with a $K$ dimension (same as the embedding vector). This filter is applied to the tweet matrix $T_i$ and performed element-wise multiplication, then the summation of all values are passed through a non-linear function to produce a new feature. Next time, this filter is again applied to the tweet matrix by moving one column towards the right and convolve with the next $h$ words of the matrix $T_i$  and passed through the non-linear function to again produce a new feature and so on. A feature $c_i$ for a window of words $W_{i:i+h-1}$ can be generated as:  

\begin{equation}
c_i = f(F~.~W_{i:i+h-1} + b)
\end{equation}
  
where, $b \in R$ is a bias and $f$ is a non-linear function. This filter is carried out to each feasible word windows having $m$ words $\{W_{1:h}, W_{2:h+1},....... ,W_{m-h+1:m}\}$ to produce a feature map.

\begin{equation}
c = [c_1, c_2, .....,c_{m-h+1}]
\end{equation}    

where $c \in R^{m-h+1}$. A simple convolution operation with a filter having size $h=3$ is represented as:

\[
\kbordermatrix{
	& W_1 & W_2 & W_3 & \dots & W_m \\
	& \alpha_{11} & \alpha_{21} & \alpha_{31} & \dots & \alpha_{m1}\\
	& \alpha_{12} & \alpha_{22} & \alpha_{32} & \dots &\alpha_{m2}\\
	& \alpha_{13} & \alpha_{23} & \alpha_{33} & \dots &\alpha_{m3}\\
	& \vdots & \vdots & \vdots & \vdots & \vdots \\
	& \alpha_{1K} & \alpha_{2K} & \alpha_{3K} & \dots & \alpha_{mK}\\
}
\bullet
\kbordermatrix{
	&  &    \\
	& f_{11} & f_{21} & f_{31}  \\
	& f_{12} & f_{22} & f_{32} \\
	& f_{13} & f_{23} & f_{33} \\
	& \vdots & \vdots & \vdots \\
	& f_{1K} & f_{2K} & f_{3K} \\
}
\]

\[\kbordermatrix{
	&      \\
	& c_1   \\
	& c_2 \\
	& c_3  \\
	& \vdots \\
	& c_{m-h+1}  \\
}
 = \kbordermatrix{\\
	& f\{\alpha_{11}f_{11} +\alpha_{21}f_{21}+\alpha_{31}f_{31}+\alpha_{12}f_{12}+ \alpha_{22}f_{22}+\alpha_{32}f_{32}+ \dots + \alpha_{1K}f_{1K} + \alpha_{2K}f_{2K}+\alpha_{3K}f_{3K}\}  \\
	& f\{\alpha_{21}f_{11} +\alpha_{31}f_{21}+\alpha_{41}f_{31}+\alpha_{22}f_{12}+ \alpha_{32}f_{22}+\alpha_{42}f_{32}+ \dots + \alpha_{2K}f_{1K} + \alpha_{3K}f_{2K}+\alpha_{4K}f_{3K}\}   \\
	& f\{\alpha_{31}f_{11} +\alpha_{41}f_{21}+\alpha_{51}f_{31}+\alpha_{32}f_{12}+ \alpha_{42}f_{22}+\alpha_{52}f_{32}+ \dots + \alpha_{3K}f_{1K} + \alpha_{4K}f_{2K}+\alpha_{5K}f_{3K}\}   \\
	& \vdots  \\
	& f\{\alpha_{(m-2)1}f_{11} +\alpha_{(m-1)1}f_{21}+\alpha_{(m)1}f_{31}+\alpha_{(m-2)2}f_{12}+ \alpha_{(m-1)2}f_{22}+\alpha_{(m)2}f_{32}+ \dots + \alpha_{(m-2)K}f_{1K} + \\ &\alpha_{(m-1)K}f_{2K}+\alpha_{(m)K}f_{3K} \}\hspace{9.4cm}  \\
}
\]

We used a Rectified Linear Unit (ReLu) \citep{nair2010rectified} as an activation function. The ReLu activation function is defined as: $f(x) = max(0,x)$, it means for $f(x) < 0$ it returns $0$ and for $f(x)\geq 0$ it returns $x$ itself. We used this function because it improves the training of CNN by speeding up the training process, as the computation step in ReLu is easier. After obtaining the feature map $c$, we applied a max-over-time pooling operation \citep{collobert2011natural} and took the maximum value from a window of size $p$, as given by equation \ref{444}.
\begin{equation}
\label{444}
\hat{c}_1 = max(c_i, c_{i+1}, c_{i+2} ,......, c_{p}), i\geq 1
\end{equation} 

The purpose of applying the pooling operation is to get the most important feature in each of the windows, i.e., one with the highest value. Similarly, we obtained a number-of-features matrix, one for each of the filters. After the convolution layer, we concatenated each of the matrices and flattened it to a single feature vector as can be seen in Figure \ref{MD}. The obtained neurons of the flattened layer are fully connected to the dense layer with sigmoid activation function, that predicts the output as the probability of occurrence of the location names in the tweet. To overcome the situation of over-fitting at the dense layers, we used dropout \citep{srivastava2014dropout} as the regularization technique. Dropout prevents the interdependency between the hidden neurons by simply dropping it out randomly with the probability of $D_p$. This allows the neural network to learn more robust features and speed up the training.  

\subsection{Representation of labels at the output layer}
The location references present in the tweet are represented to the output layer in the form of a zero-one vector. The location words are encoded as $1$ and the non-location words are encoded as $0$. For the tweets \textit{``earthquake occurred near tazeh abad kermanshah at utc earthquake tazehabad,''} \textit{``hey my friend lost everything in mexico earthquake incl his imac can you help him fix,''} and \textit{``help out give to relief for earthquake victims in kurdistan generosity via''} the location names are present at word index of $(4, 5, 6, 10), (7)$, and $(10)$ respectively. So, we put $1$ into those word indexes and the rest as $0$.

\subsection{Loss Function and Optimizer}
A loss function is used to calculate the difference between the actual and predicted values at the output layer. This loss is then back-propagated  \citep{hecht1992theory} through the output layer to adjust the weight of neurons in the network. In our case, we have a multi-labeled dataset with more than one location name, so we used binary cross entropy loss with sigmoid activation function at the output layer \citep{nam2014large} for the multi-labeled datasets. Binary cross entropy loss and sigmoid function are defined as:

\begin{eqnarray*}
Binary~cross~entropy~loss~= - \sum_{i=1}^{|L|} [y_i log(\hat{y_i}) + (1-y_i) log (1-\hat{y_i})]\\
Sigmoid~function~\sigma(x) = \frac{1}{1 + \exp(-x)}, where~ x \in R
\end{eqnarray*}

where, $|L|$ represents the total number of labels in the tweet, $y_i$, and $\hat{y_i}$ represents the actual and predicted values of the network respectively. We used Adaptive Moment Estimation (Adam) optimizer \citep{kingma2014adam} to adjust the weights by back-propagating the calculated loss.

\subsection{Hyper-parameter setting}
We define several hyper-parameters for the proposed CNN network, which can be seen in Table \ref{hyper}. In our preliminary analysis, we first experimented with the several variations of optimization algorithms, such as the Stochastic Gradient Descent (SGD), RMSProp, and Adam, by keeping all hyper-parameters constant as listed in Table \ref{hyper}. We found that Adam, with binary cross entropy, produces the lowest training loss, so we used Adam for the proposed model. Next, we experimented with the numbers 64, 128, and 256 of each filter size. The best result was found in the case of 128 filters for each filter size with the max pooling operation having a window size of 5. The proposed model was again tested with a batch size of 50, 100, and 150. A better result was found in the case of batch size 50. We tested the model with a dropout value of 0.2, 0.3, and 0.5; the performance of the model was better in the case of the 0.2 dropout. Similarly, the system was tested by varying the epoch sizes; the performance of the model did not affect as much after 100 epochs, so we fixed the number of epochs to 100 for all our other experiments.

\begin{table}
	\caption{Description of the Hyper-parameters}
	\begin{center}
		\label{hyper}
		\begin{tabular}{ |p{4cm}|p{3cm}|}
			\hline
			Description &	Values   \\ \hline
			Filter region size & 2,3,4 \\ \hline
			Feature maps & 128 \\ \hline
			Pooling window size  & 5 \\ \hline
			Pooling & Max pooling \\ \hline
			Activation function & ReLu \\ \hline
			Dense layer & 60 neurons \\ \hline
			Dropout rate & 0.2 \\ \hline
			Learning rate & 0.001 \\ \hline
			Batch size & 50 \\ \hline
			Epochs & 100 \\ \hline
		\end{tabular}
	\end{center}\
\end{table}

\section{Result}
\label{Res}
We performed several experiments to evaluate our proposed model and extract location words from the earthquake-related tweets. To minimize the bias, we used 10-fold cross validation \citep{kohavi1995study}. It is a technique to randomly partition the data sample into ten equal subsamples in which one subsample is used to validate the system, whereas the remaining nine subsamples are used to train the model. This process is repeated ten times, with each of the ten subsamples used just once as the validation data. The results from each of the folds are averaged to estimate the overall system performance. According to our observation, in most of the tweets, the number of words is 60 at the most. Hence, we used 60 neurons at the input layer to represent the words of each tweet and 60 neurons at the output layer to encode the presence or absence of location references.

\subsection{Evaluation metrics}
To evaluate the proposed model, we used Precision, Recall, $F_1$-score, the Hamming loss, the Jaccard similarity, and the Exact matching score. These metrics are widely used in the case of a multi-labeled dataset \citep{charte2015working}. Say a multi-labeled dataset contains a total of $N$ instances; each instance $N_i$ can be represented as $(x_i, y_i)$, where $x_i$ is the set of attributes. $ y_i \subseteq L$ is the set of labels, where $L$ represents the total number of labels used in the dataset. Suppose $y_i$ and $\hat{y}_i$ represents the subset of true and predicted labels respectively for the $i^{th}$ instance, then the metrics can be described for the $i^{th}$ instance by the given formulae. 

\begin{itemize}
\item \textbf{Precision:} This is the number of accurately predicted location words to the total number of predicted location words. It is computed as given in equation \ref{precision}. The range of precision varies between 0 and 1, where 1 is the best and 0 is the worst value.
\begin{equation}
\label{precision}
\mbox{Precision} = \frac{\mbox{Number of accurately predicted location words}}{\mbox{Total number of predicted location words}} = \frac{\mid y_i \cap \hat{y}_i \mid }{\mid \hat{y}_i \mid}
\end{equation}

\item \textbf{Recall:} This is the number of accurately predicted location words to the total number of actual location words in the tweet. It is computed as given in equation \ref{Recall}.  The range of recall varies between 0 and 1, where 1 is the best and 0 is the worst value.
\begin{equation}
\label{Recall}
\mbox{Recall} =\frac{\mbox{Number of accurately predicted location words}}{\mbox{Total number of actual location words}} = \frac{\mid y_i \cap \hat{y}_i \mid }{\mid y_i \mid}
\end{equation}

\item \textbf{$F_1$-score:} This is the harmonic mean between Precision and Recall, which gives the balanced evaluation between them. It can be represented by equation \ref{Fmeasure}. The range of $F_1$-score varies between 0 and 1, where 1 is the best and 0 is the worst value. 
\begin{equation}
\label{Fmeasure}
\mbox{$F_1$-score} = 2 \times \frac{\mbox{Precision} \times \mbox{Recall}}{\mbox{Precision + Recall}}
\end{equation}

\item \textbf{Hamming Loss:} This is the number of the wrong predictions to the total number of predictions. It is calculated by equation \ref{hamming}. The indicator function $1( y_i = \hat{y}_i)$ returns 1 when the expression is true, otherwise it returns to 0. The range of Hamming loss varies between 0 and 1, where 0 is the best and 1 is the worst value.
\begin{equation}
\label{hamming}
\mbox{Hamming Loss} =\frac{\mbox{Number of wrong prediction}}{\mbox{Total number of prediction}} =\frac{1}{{\mid L \mid}} \displaystyle\sum_{j=1}^{\mid L \mid} 1(y_{ij} \neq \hat{y}_{ij}) 
\end{equation}

\item \textbf{Jaccard similarity:} This is the number of accurately predicted location words to the union of actual and predicted location words. It is represented in equation \ref{jaccard}. The range of the Jaccard index varies between 0 and 1, where 1 is the best and 0 is the worst value. 
\begin{equation}
\label{jaccard}
 \mbox{Jaccard similarity} = \frac{\mbox{Number of accurately predicted location words} }{\mbox{Union of actual \& predicted location words}}= \frac{\mid y_i \cap \hat{y}_i \mid}{\mid y_i \cup \hat{y}_i \mid} 
\end{equation}

\item \textbf{Exact Matching score:} This can be computed by equation \ref{exact}. The indicator function $1( y_i = \hat{y}_i)$ returns to 1 if all the predicted location and non-location words are as true as the actual one, otherwise it returns to 0.  
\begin{equation}
\label{exact}
\mbox{Exact Matching score} = 1( y_i = \hat{y}_i) 
\end{equation}
\end{itemize}
\begin{table}[t]
	\centering
	\caption{Result of the 2-CNN and 2-Dense with dropout model with combinations of 2-, 3-, 4-, and 5-gram filter size}
	\label{2cnn2densedrop}
	\begin{tabular}{|l|l|l|l|l|p{1.5cm}|p{1.5cm}|p{1.5cm}|}
		\hline
		Approach & Filter size & Precision & Recall & F1-score & Hamming loss & Jaccard similarity & Exact matching \\ \cline{1-8} 
		& 2           & 0.57      & 0.50    & 0.52     & 0.018        & 0.531              & 0.429          \\ \cline{2-8} 
		& 3           & 0.60      & 0.45   & 0.50      & 0.016        & 0.551              & 0.473          \\ \cline{2-8} 
		& 4           & 0.62      & 0.47   & 0.52     & 0.015        & 0.559              & 0.475          \\ \cline{2-8} 
		& 5           & 0.65      & 0.47   & 0.53     & 0.015        & 0.568              & 0.484          \\ \cline{2-8} 
		& 2,3         & 0.97      & 0.86   & 0.90      & 0.003        & 0.900                & 0.849          \\ \cline{2-8} 
		& 2,4         & 0.95      & 0.89   & 0.91     & 0.003        & 0.924              & 0.886          \\ \cline{2-8} 
		& 2,5         & 0.96      & 0.92   & 0.94     & 0.002        & 0.948              & 0.918          \\ \cline{2-8} 
		2-CNN+2-Dense+Dropout & 3,4         & 0.97      & 0.89   & 0.92     & 0.003        & 0.914              & 0.875          \\ \cline{2-8} 
		& 3,5         & 0.98      & 0.93   & 0.95     & 0.002        & 0.949             & 0.922          \\ \cline{2-8} 
		& 4,5         & 0.97      & 0.90    & 0.92     & 0.003        & 0.926              & 0.890           \\ \cline{2-8} 
		& \textbf{2,3,4}       & \textbf{0.97}      & \textbf{0.93}   & \textbf{0.95}     & \textbf{0.002}        & \textbf{0.953}              & \textbf{0.924}          \\ \cline{2-8} 
		& 2,3,5       & 0.98      & 0.92   & 0.94     & 0.002        & 0.938              & 0.906          \\ \cline{2-8} 
		& 2,4,5       & 0.98      & 0.91   & 0.94     & 0.002        & 0.949              & 0.925          \\ \cline{2-8} 
		& 3,4,5       & 0.98 & 0.90 & 0.93 & 0.002 & 0.926 & 0.892  \\ \cline{2-8} 
		& 2,3,4,5     & 0.99      & 0.91   & 0.94     & 0.002        & 0.934              & 0.892          \\ \hline
	\end{tabular}
\end{table}

Say there is a tweet that says \textit{``very strong earthquake felt here, kermadec island, new zealand"}. This tweet has four location words \textit{kermadec}, \textit{island}, \textit{new}, and \textit{zealand} occurring at word positions 6, 7, 8, and 9 respectively. So the real output can be encoded as $y_i$ = [0 0 0 0 0 1 1 1 1]. In case 1, our system predicted the output $\hat{y}_i$ = [0 0 0 0 1 0 0 1 1]. In this prediction, two location words \textit{new}, and \textit{zealand} are predicted correctly, while two location words \textit{kermadec}, and \textit{island} are wrongly predicted as non-location words. One non-location word \textit{here} was wrongly predicted as a location word. So, from the definition of evaluation metrics, precision = number of accurately predicted location words [at position (8, 9)]/number of predicted location words [at position (5, 8, 9)] = 2/3 = 0.66, recall = number of accurately predicted location words [at position (8, 9)]/number of actual location words [at position (6, 7, 8, 9)] = 2/4 = 0.5, $F_1$-score = harmonic mean of precision and recall = $2\times(0.66\times0.5)/(0.66+0.5)$ = 0.57, hamming loss = number of wrong prediction [at position (5, 6, 7)]/ total number of prediction = 3/9 = 0.33, Jaccard similarity = number of accurately predicted location words [at position (8, 9)]/union of actual and predicted location words [at position (5, 6, 7, 8, 9)] = 2/5 = 0.4, Exact matching score = 0 (as the total location and non-location words are not correctly predicted). In case 2, if system predicted the following output $\hat{y}_i$ = [0 0 0 0 0 1 1 1 1], means all location and non-location words correctly predicted. Then, precision = 3/3 = 1.0, recall = 4/4 = 1.0, $F_1$-score = $2\times(1.0\times1.0)/(1.0+1.0)$ = 1.0, hamming loss = 9/9 = 1.0, Jaccard similarity = 4/4 = 1.0, Exact matching score = 1.0 (as location and non-location words are correctly predicted).


\subsubsection{Filter size estimation}
Since the tweet contains more than one location word, determining the suitable filter size is a primary concern. We started experimentation with the different models: (i) 1-CNN + 2-Dense; (ii) 1-CNN + 2-Dense with Dropout; (iii) 2-CNN + 2-Dense; and (iv) 2-CNN + 2-Dense with Dropout. We used different combinations of 2-gram, 3-gram, 4-gram, and 5-gram filters with each of the models to extract the features from the tweet. The best-performing model was found to be 2-CNN + 2-Dense with dropout. The results of this model for different filter sizes are listed in Table \ref{2cnn2densedrop}. In the analysis, we found that the use of a single filter size was not adequate as all the models performed very badly. It can be seen from Table \ref{2cnn2densedrop}, that the use of individual 2-gram, 3-gram, 4-gram, and 5-gram filters did not perform well as it yielded the $F_1$-score of 0.52, 0.50, 0.52, and 0.53 respectively. The use of the two filters together performed better than the individual filters. The best result was achieved when we used 2-gram, 3-gram, and 4-gram filters together, which yielded the precision of 0.97, a recall of 0.93, an $F_1$-score of 0.95, a hamming loss of 0.002, a Jaccard similarity of 0.953, and an exact matching score of 0.924. So, for all further experiments, we fixed the filter sizes of 2-gram, 3-gram, and 4-gram and used them together.

\subsubsection{Effect of varying the dense layers}
To determine the suitable number of dense layers, we experimented with ten different combinations of CNN, dense, and dropout layers. The results of different combinations are tabulated in Table \ref{finalresult}. As can be seen from Table \ref{finalresult}, two dense layers yielded the best performance metrics with 2-CNN + 2-Dense + Dropout model. So for further experimentation, we fixed the number of dense layers to two. 

\begin{table}[h]
	\centering
	\caption{Result of the proposed Convolutional Neural Network- (CNN) based model with filter size of 2-gram, 3-gram, and 4-gram.}
	\label{finalresult}
	\begin{tabular}{ | l | P{1.3cm} | P{1.3cm} | P{1.3cm} |P{1.3cm}| P{1.5cm} |P{1.4cm}|}
		
		\hline
		Approach & Precision & Recall & $F_1$-score & Hamming Loss & Jaccard Similarity & Exact matching\\ \hline

		1-CNN + 1-Dense (Baseline) & 0.83  & 0.77 & 0.79 & 0.005 & 0.843 & 0.800 \\ \hline     
		
		1-CNN + 2-Dense & 0.85 & 0.76 & 0.79 & 0.007 & 0.815 & 0.751 \\ \hline
		
		1-CNN + 2-Dense +Dropout & 0.91 & 0.83 & 0.86 & 0.004 & 0.886 & 0.840 \\ \hline
		
		1-CNN + 3-Dense & 0.89 & 0.81 & 0.84 & 0.005 & 0.860 & 0.810 \\ \hline
		
		1-CNN + 3-Dense + Dropout & 0.91 & 0.84 & 0.86 & 0.004 & 0.887 & 0.859 \\ \hline

		2-CNN + 1-Dense & 0.95 & 0.88 & 0.90 & 0.003 & 0.906 & 0.849 \\ \hline
		
		2-CNN + 2-Dense & 0.94 & 0.86 & 0.89 & 0.003 & 0.898 & 0.843 \\ \hline
		
		\textbf{2-CNN + 2-Dense + Dropout} & \textbf{0.97} & \textbf{0.93} & \textbf{0.95} & \textbf{0.002} & \textbf{0.953} & \textbf{0.924} \\ \hline
		
		2-CNN + 3-Dense & 0.95 & 0.87 & 0.90 & 0.003 & 0.899 & 0.849 \\ \hline
		
		2-CNN + 3-Dense + Dropout & 0.96 & 0.90 & 0.92 & 0.003 & 0.933 & 0.900 \\ \hline	
	\end{tabular}
\end{table}

\subsubsection{Effect of varying the CNN layers}
The number of CNN layers was the most important parameter to decide, as the addition of one more CNN layer in 1-CNN outperformed all previous 1-CNN models. So, we did further experimentation by adding more CNN layers in the model. In order to do that, we developed models starting from 1-CNN to 4-CNN with two dense layers and filter sizes of 2-grams, 3-grams, and 4-grams together. The result of 1-CNN, 2-CNN, 3-CNN, and 4-CNN with the combination of 2-Dense and dropout are listed in Table \ref{cnn_comp}. We found that the use of 3-CNN yielded the best result. Our final model can be summarized as: (i) Number of CNN layers = 3; (ii) Filter size = 2, 3, 4; (iii) Number of dense layers = 2; and (iv) and Dropout = 0.2.

\begin{table}[h]
	\centering
	\caption{Result of the different combinations of CNN layers with filter sizes of 2-grams, 3-grams, and 4-grams}
	\label{cnn_comp}
	\begin{tabular}{| l | P{1.3cm} | P{1.3cm} | P{1.3cm} |P{1.3cm}| P{1.5cm} |P{1.4cm}|}
		
		\hline
		Approach & Precision & Recall & $F_1$-score & Hamming Loss & Jaccard Similarity & Exact matching\\ \hline 
		
		1-CNN + 2-Dense +Dropout & 0.91 & 0.83 & 0.86 & 0.004 & 0.886 & 0.840 \\ \hline

		2-CNN + 2-Dense + Dropout & 0.97 & 0.93 & 0.95 & 0.002 & 0.953 & 0.924 \\ \hline
		
		\textbf{3-CNN + 2-Dense + Dropout} & \textbf{0.98} & \textbf{0.95} & \textbf{0.96} & \textbf{0.002} & \textbf{0.962} & \textbf{0.929} \\ \hline
		
		4-CNN + 2-Dense + Dropout & 0.98 & 0.94 & 0.96 & 0.002 & 0.956 & 0.927 \\ \hline	
	\end{tabular}
\end{table}

\section{Discussion and Implications}
\label{Dis}
In this work, we proposed a Convolutional Neural Network- (CNN) based model that learns the salient features from tweets to predict the location references mentioned in it. The proposed CNN-based models could extract the location references from the tweets with significant accuracy. In our case, the proposed CNN-based model can find the location information of almost every granularity, such as streets, buildings, the city, district, and country name with very significant accuracy. The use of 2-gram, 3-gram, and 4-gram filters together with the 3-CNN and 2-Dense with a dropout performed best with the precision of 0.98, a recall of 0.95, an $F_1$-score of 0.96, a hamming loss of 0.002, a Jaccard similarity of 0.962, and the exact matching score of 0.929. The precision of 0.98 means we can extract 98 true location-referring words out of 100 location word predictions. The recall of 0.95 means we can identify 95 location-referring words out of the total 100 real location references contained in the tweet. The use of n-gram features played a major role in finding location references. The proposed system did not perform well when we applied 2-gram, 3-gram, 4-gram, and 5-gram filters individually. The use of 2-gram, 3-gram, and 4-gram together performed best, because the several location-referring names have more than one word in a tweet, as can be seen in Table \ref{dataset}. As a single location name can have more than one words to represent it, finding all the words that refer to a single location is very important. In many cases, individual words of a single location name do not preserve any meaning in terms of location. For example, the tweet \textit{``News Moderate earthquake 5.8 mag, 91 km S of Raoul Island, New Zealand - BREAKING http://ow.ly/oFAE50ga6z2"} had four location references: \textit{Raoul, Island, New, and Zealand}. If the system cannot predict all four location references together, then it is of no use as the individual words \textit{Raoul, Island, New, and Zealand} do not have any meaning in terms of location reference. In the analysis, we also found that some of the tweets have five or more location words, but most of the times they were not consecutive. They were mostly with \# sign and were distributed across the whole text. This can be one of the reasons why the use of 5-gram or more than 5-gram filter with 2-gram, 3-gram, and 4-gram did not improve the performance. We used the exact matching metric to measure the system performance. The exact matching was set to true if all the predicted location-referring word and non-location-referring words of a tweet are totally matched with the actual location-referring words and non-location-referring word respectively. Although the exact matching is a very strict evaluation metric, the proposed system can identify all the location references with the 92.9\% exact match.  

Some similar works have been reported for location word extraction from tweets. However, due to the inaccessibility of their dataset, we cannot directly compare our results with them. Hence, the comparison is merely in terms of the accuracy achieved over their datasets. \cite{malmasi2015location} used 3,000 tweets related to general topics and found an $F_1$-score of 0.792. Similarly, \cite{gelernter2013algorithm} used 3,987 tweets related to earthquakes and found an $F_1$-score of 0.90, and \cite{lingad2013location} used 2,878 tweets related to several disasters, such as the Queensland flood 2012, the Christchurch earthquake 2012, and the England riots of 2011 and achieved an $F_1$-score of 0.902. In line with their studies, our system, which was validated by 5,107 tweets, achieved an $F_1$-score of 0.96 in finding location references mentioned in the tweets. Some other works \citep{nguyen2017real, perol2018convolutional} also utilized convolutional neural networks in case of the earthquake, but their objectives were different. \cite{nguyen2017real} used CNN to classify tweets into relevant and non-relevant classes and proposed an algorithm for timely detection of the earthquake. \cite{perol2018convolutional} used CNN to detect the earthquake from seismic waves. The proposed convolutional neural network-based model is independent of manual feature engineering. This system works well even in the situation of noisy tweet texts. We preserved the privacy of users during this work as we only used tweet text and removed all other metadata and mentioned user names in the tweets. The main theoretical implication of this work is in the development of models based on the convolutional neural network for geographical location estimation without going through POS tagging as when using conventional Named Entity Recognition tools. The major practical implication of this work is in: (i) early event localization; (ii) finding the location of victims; and (iii) the rescue and relief operation. Irrespective of this, the proposed system can be utilized in case of civil unrest, targeted advertising, observing regional human behavior, real-time road traffic management, and in various location-based services. The proposed system can be easily integrated with event detection models.

\section{Conclusion}
\label{Con}
The extraction of location information from the tweets is a challenging task as tweets have various noise in terms of grammatical mistakes, spelling mistakes, and non-standard abbreviations. We have proposed a convolutional neural network-based model for finding location references present in the tweets. We used earthquake-related tweets and performed our implementation with several configurations of convolution layers with the dense layers. We achieved our best result with an $F_1$-score of 0.96 when we used 3-CNN and 2-Dense layers with dropout. We can find location information of several granularities, such as streets, buildings, city, district, and country name with very impressive accuracy. This system can be utilized in other domain, such as road traffic management and sports events, in several location-based services by training it with domain-specific tweets. The trained models can even be integrated with mobile devices to also find location information of tweets on the fly. The limitation of this work is related to the manual labeling of location references in the tweets, as it requires huge human effort. A semi-supervised approach can be used to reduce the task of manual labeling to some extent. In this work, we only considered English language tweets, but users also post tweets in their regional languages during a crisis. So, a deep neural network-based model can be created to deal with the multi-linguality issues of tweets.  



\section*{References}

\begin{thebibliography}{86}
\expandafter\ifx\csname natexlab\endcsname\relax\def\natexlab#1{#1}\fi
\providecommand{\url}[1]{\texttt{#1}}
\providecommand{\href}[2]{#2}
\providecommand{\path}[1]{#1}
\providecommand{\DOIprefix}{doi:}
\providecommand{\ArXivprefix}{arXiv:}
\providecommand{\URLprefix}{URL: }
\providecommand{\Pubmedprefix}{pmid:}
\providecommand{\doi}[1]{\href{http://dx.doi.org/#1}{\path{#1}}}
\providecommand{\Pubmed}[1]{\href{pmid:#1}{\path{#1}}}
\providecommand{\bibinfo}[2]{#2}
\ifx\xfnm\relax \def\xfnm[#1]{\unskip,\space#1}\fi
\bibitem[{Ajao et~al.(2015)Ajao, Hong \& Liu}]{ajao2015survey}
\bibinfo{author}{Ajao, O.}, \bibinfo{author}{Hong, J.}, \&
  \bibinfo{author}{Liu, W.} (\bibinfo{year}{2015}).
\newblock \bibinfo{title}{A survey of location inference techniques on
  twitter}.
\newblock {\it \bibinfo{journal}{Journal of Information Science}\/},  {\it
  \bibinfo{volume}{41}\/}, \bibinfo{pages}{855--864}.
\bibitem[{Al-Olimat et~al.(2017)Al-Olimat, Thirunarayan, Shalin \&
  Sheth}]{al2017location}
\bibinfo{author}{Al-Olimat, H.~S.}, \bibinfo{author}{Thirunarayan, K.},
  \bibinfo{author}{Shalin, V.}, \& \bibinfo{author}{Sheth, A.}
  (\bibinfo{year}{2017}).
\newblock \bibinfo{title}{Location name extraction from targeted text streams
  using gazetteer-based statistical language models}.
\newblock {\it \bibinfo{journal}{arXiv preprint arXiv:1708.03105}\/}, .
\bibitem[{Alexander(2014)}]{alexander2014social}
\bibinfo{author}{Alexander, D.~E.} (\bibinfo{year}{2014}).
\newblock \bibinfo{title}{Social media in disaster risk reduction and crisis
  management}.
\newblock {\it \bibinfo{journal}{Science and engineering ethics}\/},  {\it
  \bibinfo{volume}{20}\/}, \bibinfo{pages}{717--733}.
\bibitem[{Atefeh \& Khreich(2015)}]{atefeh2015survey}
\bibinfo{author}{Atefeh, F.}, \& \bibinfo{author}{Khreich, W.}
  (\bibinfo{year}{2015}).
\newblock \bibinfo{title}{A survey of techniques for event detection in
  twitter}.
\newblock {\it \bibinfo{journal}{Computational Intelligence}\/},  {\it
  \bibinfo{volume}{31}\/}, \bibinfo{pages}{132--164}.
\bibitem[{Blei et~al.(2003)Blei, Ng \& Jordan}]{blei2003latent}
\bibinfo{author}{Blei, D.~M.}, \bibinfo{author}{Ng, A.~Y.}, \&
  \bibinfo{author}{Jordan, M.~I.} (\bibinfo{year}{2003}).
\newblock \bibinfo{title}{Latent dirichlet allocation}.
\newblock {\it \bibinfo{journal}{Journal of machine Learning research}\/},
  {\it \bibinfo{volume}{3}\/}, \bibinfo{pages}{993--1022}.
\bibitem[{Charte \& Charte(2015)}]{charte2015working}
\bibinfo{author}{Charte, F.}, \& \bibinfo{author}{Charte, D.}
  (\bibinfo{year}{2015}).
\newblock \bibinfo{title}{Working with multilabel datasets in r: the mldr
  package}.
\newblock {\it \bibinfo{journal}{R J}\/},  {\it \bibinfo{volume}{7}\/},
  \bibinfo{pages}{149--162}.
\bibitem[{Cheng et~al.(2010)Cheng, Caverlee \& Lee}]{cheng2010you}
\bibinfo{author}{Cheng, Z.}, \bibinfo{author}{Caverlee, J.}, \&
  \bibinfo{author}{Lee, K.} (\bibinfo{year}{2010}).
\newblock \bibinfo{title}{You are where you tweet: a content-based approach to
  geo-locating twitter users}.
\newblock In {\it \bibinfo{booktitle}{Proceedings of the 19th ACM international
  conference on Information and knowledge management}\/} (pp.
  \bibinfo{pages}{759--768}).
\newblock \bibinfo{organization}{ACM}.
\bibitem[{Chiu \& Nichols(2015)}]{chiu2015named}
\bibinfo{author}{Chiu, J.~P.}, \& \bibinfo{author}{Nichols, E.}
  (\bibinfo{year}{2015}).
\newblock \bibinfo{title}{Named entity recognition with bidirectional
  lstm-cnns}.
\newblock {\it \bibinfo{journal}{arXiv preprint arXiv:1511.08308}\/}, .
\bibitem[{Chowdhury et~al.(2013)Chowdhury, Imran, Asghar, Amer-Yahia \&
  Castillo}]{chowdhury2013tweet4act}
\bibinfo{author}{Chowdhury, S.~R.}, \bibinfo{author}{Imran, M.},
  \bibinfo{author}{Asghar, M.~R.}, \bibinfo{author}{Amer-Yahia, S.}, \&
  \bibinfo{author}{Castillo, C.} (\bibinfo{year}{2013}).
\newblock \bibinfo{title}{Tweet4act: Using incident-specific profiles for
  classifying crisis-related messages.}
\newblock In {\it \bibinfo{booktitle}{ISCRAM}\/}.
\newblock \bibinfo{organization}{Citeseer}.
\bibitem[{Collobert et~al.(2011)Collobert, Weston, Bottou, Karlen, Kavukcuoglu
  \& Kuksa}]{collobert2011natural}
\bibinfo{author}{Collobert, R.}, \bibinfo{author}{Weston, J.},
  \bibinfo{author}{Bottou, L.}, \bibinfo{author}{Karlen, M.},
  \bibinfo{author}{Kavukcuoglu, K.}, \& \bibinfo{author}{Kuksa, P.}
  (\bibinfo{year}{2011}).
\newblock \bibinfo{title}{Natural language processing (almost) from scratch}.
\newblock {\it \bibinfo{journal}{Journal of Machine Learning Research}\/},
  {\it \bibinfo{volume}{12}\/}, \bibinfo{pages}{2493--2537}.
\bibitem[{D{\"a}niken \& Cieliebak(2017)}]{von2017transfer}
\bibinfo{author}{D{\"a}niken, P.}, \& \bibinfo{author}{Cieliebak, M.}
  (\bibinfo{year}{2017}).
\newblock \bibinfo{title}{Transfer learning and sentence level features for
  named entity recognition on tweets}.
\newblock In {\it \bibinfo{booktitle}{Proceedings of the 3rd Workshop on Noisy
  User-generated Text}\/} (pp. \bibinfo{pages}{166--171}).
\bibitem[{Do et~al.(2017)Do, Nguyen, Tsiligianni, Cornelis \&
  Deligiannis}]{do2017multiview}
\bibinfo{author}{Do, T.~H.}, \bibinfo{author}{Nguyen, D.~M.},
  \bibinfo{author}{Tsiligianni, E.}, \bibinfo{author}{Cornelis, B.}, \&
  \bibinfo{author}{Deligiannis, N.} (\bibinfo{year}{2017}).
\newblock \bibinfo{title}{Multiview deep learning for predicting twitter users'
  location}.
\newblock {\it \bibinfo{journal}{arXiv preprint arXiv:1712.08091}\/}, .
\bibitem[{Dutt et~al.(2018)Dutt, Hiware, Ghosh \& Bhaskaran}]{dutt2018savitr}
\bibinfo{author}{Dutt, R.}, \bibinfo{author}{Hiware, K.},
  \bibinfo{author}{Ghosh, A.}, \& \bibinfo{author}{Bhaskaran, R.}
  (\bibinfo{year}{2018}).
\newblock \bibinfo{title}{Savitr: A system for real-time location extraction
  from microblogs during emergencies}.
\newblock {\it \bibinfo{journal}{arXiv preprint arXiv:1801.07757}\/}, .
\bibitem[{Gelernter \& Balaji(2013)}]{gelernter2013algorithm}
\bibinfo{author}{Gelernter, J.}, \& \bibinfo{author}{Balaji, S.}
  (\bibinfo{year}{2013}).
\newblock \bibinfo{title}{An algorithm for local geoparsing of microtext}.
\newblock {\it \bibinfo{journal}{GeoInformatica}\/},  {\it
  \bibinfo{volume}{17}\/}, \bibinfo{pages}{635--667}.
\bibitem[{Gelernter \& Mushegian(2011)}]{gelernter2011geo}
\bibinfo{author}{Gelernter, J.}, \& \bibinfo{author}{Mushegian, N.}
  (\bibinfo{year}{2011}).
\newblock \bibinfo{title}{Geo-parsing messages from microtext}.
\newblock {\it \bibinfo{journal}{Transactions in GIS}\/},  {\it
  \bibinfo{volume}{15}\/}, \bibinfo{pages}{753--773}.
\bibitem[{Giridhar et~al.(2015)Giridhar, Abdelzaher, George \&
  Kaplan}]{giridhar2015quality}
\bibinfo{author}{Giridhar, P.}, \bibinfo{author}{Abdelzaher, T.},
  \bibinfo{author}{George, J.}, \& \bibinfo{author}{Kaplan, L.}
  (\bibinfo{year}{2015}).
\newblock \bibinfo{title}{On quality of event localization from social network
  feeds}.
\newblock In {\it \bibinfo{booktitle}{Pervasive Computing and Communication
  Workshops (PerCom Workshops), 2015 IEEE International Conference on}\/} (pp.
  \bibinfo{pages}{75--80}).
\newblock \bibinfo{organization}{IEEE}.
\bibitem[{Goldberg(2016)}]{goldberg2016primer}
\bibinfo{author}{Goldberg, Y.} (\bibinfo{year}{2016}).
\newblock \bibinfo{title}{A primer on neural network models for natural
  language processing.}
\newblock {\it \bibinfo{journal}{J. Artif. Intell. Res.(JAIR)}\/},  {\it
  \bibinfo{volume}{57}\/}, \bibinfo{pages}{345--420}.
\bibitem[{Gu et~al.(2016)Gu, Qian \& Chen}]{gu2016twitter}
\bibinfo{author}{Gu, Y.}, \bibinfo{author}{Qian, Z.~S.}, \&
  \bibinfo{author}{Chen, F.} (\bibinfo{year}{2016}).
\newblock \bibinfo{title}{From twitter to detector: Real-time traffic incident
  detection using social media data}.
\newblock {\it \bibinfo{journal}{Transportation research part C: emerging
  technologies}\/},  {\it \bibinfo{volume}{67}\/}, \bibinfo{pages}{321--342}.
\bibitem[{Hecht et~al.(2011)Hecht, Hong, Suh \& Chi}]{hecht2011tweets}
\bibinfo{author}{Hecht, B.}, \bibinfo{author}{Hong, L.}, \bibinfo{author}{Suh,
  B.}, \& \bibinfo{author}{Chi, E.~H.} (\bibinfo{year}{2011}).
\newblock \bibinfo{title}{Tweets from justin bieber's heart: the dynamics of
  the location field in user profiles}.
\newblock In {\it \bibinfo{booktitle}{Proceedings of the SIGCHI conference on
  human factors in computing systems}\/} (pp. \bibinfo{pages}{237--246}).
\newblock \bibinfo{organization}{ACM}.
\bibitem[{Hecht-Nielsen(1992)}]{hecht1992theory}
\bibinfo{author}{Hecht-Nielsen, R.} (\bibinfo{year}{1992}).
\newblock \bibinfo{title}{Theory of the backpropagation neural network}.
\newblock In {\it \bibinfo{booktitle}{Neural networks for perception}\/} (pp.
  \bibinfo{pages}{65--93}).
\newblock \bibinfo{publisher}{Elsevier}.
\bibitem[{Hoang et~al.(2016)Hoang, Cher, Prasetyo \&
  Lim}]{hoang2016crowdsensing}
\bibinfo{author}{Hoang, T.}, \bibinfo{author}{Cher, P.~H.},
  \bibinfo{author}{Prasetyo, P.~K.}, \& \bibinfo{author}{Lim, E.-P.}
  (\bibinfo{year}{2016}).
\newblock \bibinfo{title}{Crowdsensing and analyzing micro-event tweets for
  public transportation insights}.
\newblock In {\it \bibinfo{booktitle}{2016 IEEE International Conference on Big
  Data (Big Data)}\/} (pp. \bibinfo{pages}{2157--2166}).
\newblock \bibinfo{organization}{IEEE}.
\bibitem[{Hofmann(1999)}]{hofmann1999probabilistic}
\bibinfo{author}{Hofmann, T.} (\bibinfo{year}{1999}).
\newblock \bibinfo{title}{Probabilistic latent semantic analysis}.
\newblock In {\it \bibinfo{booktitle}{Proceedings of the Fifteenth conference
  on Uncertainty in artificial intelligence}\/} (pp.
  \bibinfo{pages}{289--296}).
\newblock \bibinfo{organization}{Morgan Kaufmann Publishers Inc.}
\bibitem[{Huang et~al.(2014)Huang, Cao \& Wang}]{huang2014tweets}
\bibinfo{author}{Huang, Q.}, \bibinfo{author}{Cao, G.}, \&
  \bibinfo{author}{Wang, C.} (\bibinfo{year}{2014}).
\newblock \bibinfo{title}{From where do tweets originate?: a gis approach for
  user location inference}.
\newblock In {\it \bibinfo{booktitle}{Proceedings of the 7th ACM SIGSPATIAL
  International Workshop on Location-Based Social Networks}\/} (pp.
  \bibinfo{pages}{1--8}).
\newblock \bibinfo{organization}{ACM}.
\bibitem[{Huang et~al.(2015)Huang, Xu \& Yu}]{huang2015bidirectional}
\bibinfo{author}{Huang, Z.}, \bibinfo{author}{Xu, W.}, \& \bibinfo{author}{Yu,
  K.} (\bibinfo{year}{2015}).
\newblock \bibinfo{title}{Bidirectional lstm-crf models for sequence tagging}.
\newblock {\it \bibinfo{journal}{arXiv preprint arXiv:1508.01991}\/}, .
\bibitem[{Ikawa et~al.(2012)Ikawa, Enoki \& Tatsubori}]{ikawa2012location}
\bibinfo{author}{Ikawa, Y.}, \bibinfo{author}{Enoki, M.}, \&
  \bibinfo{author}{Tatsubori, M.} (\bibinfo{year}{2012}).
\newblock \bibinfo{title}{Location inference using microblog messages}.
\newblock In {\it \bibinfo{booktitle}{Proceedings of the 21st International
  Conference on World Wide Web}\/} (pp. \bibinfo{pages}{687--690}).
\newblock \bibinfo{organization}{ACM}.
\bibitem[{Imran et~al.(2015)Imran, Castillo, Diaz \&
  Vieweg}]{imran2015processing}
\bibinfo{author}{Imran, M.}, \bibinfo{author}{Castillo, C.},
  \bibinfo{author}{Diaz, F.}, \& \bibinfo{author}{Vieweg, S.}
  (\bibinfo{year}{2015}).
\newblock \bibinfo{title}{Processing social media messages in mass emergency: A
  survey}.
\newblock {\it \bibinfo{journal}{ACM Computing Surveys (CSUR)}\/},  {\it
  \bibinfo{volume}{47}\/}, \bibinfo{pages}{67}.
\bibitem[{Imran et~al.(2014{\natexlab{a}})Imran, Castillo, Lucas, Meier \&
  Rogstadius}]{imran2014coordinating}
\bibinfo{author}{Imran, M.}, \bibinfo{author}{Castillo, C.},
  \bibinfo{author}{Lucas, J.}, \bibinfo{author}{Meier, P.}, \&
  \bibinfo{author}{Rogstadius, J.} (\bibinfo{year}{2014}{\natexlab{a}}).
\newblock \bibinfo{title}{Coordinating human and machine intelligence to
  classify microblog communications in crises.}
\newblock In {\it \bibinfo{booktitle}{ISCRAM}\/}.
\bibitem[{Imran et~al.(2014{\natexlab{b}})Imran, Castillo, Lucas, Meier \&
  Vieweg}]{imran2014aidr}
\bibinfo{author}{Imran, M.}, \bibinfo{author}{Castillo, C.},
  \bibinfo{author}{Lucas, J.}, \bibinfo{author}{Meier, P.}, \&
  \bibinfo{author}{Vieweg, S.} (\bibinfo{year}{2014}{\natexlab{b}}).
\newblock \bibinfo{title}{Aidr: Artificial intelligence for disaster response}.
\newblock In {\it \bibinfo{booktitle}{Proceedings of the 23rd International
  Conference on World Wide Web}\/} (pp. \bibinfo{pages}{159--162}).
\newblock \bibinfo{organization}{ACM}.
\bibitem[{Imran et~al.(2013)Imran, Elbassuoni, Castillo, Diaz \&
  Meier}]{imran2013extracting}
\bibinfo{author}{Imran, M.}, \bibinfo{author}{Elbassuoni, S.},
  \bibinfo{author}{Castillo, C.}, \bibinfo{author}{Diaz, F.}, \&
  \bibinfo{author}{Meier, P.} (\bibinfo{year}{2013}).
\newblock \bibinfo{title}{Extracting information nuggets from disaster-related
  messages in social media.}
\newblock In {\it \bibinfo{booktitle}{Iscram}\/}.
\bibitem[{Itoh et~al.(2016)Itoh, Yoshinaga \& Toyoda}]{itoh2016spatio}
\bibinfo{author}{Itoh, M.}, \bibinfo{author}{Yoshinaga, N.}, \&
  \bibinfo{author}{Toyoda, M.} (\bibinfo{year}{2016}).
\newblock \bibinfo{title}{Spatio-temporal event visualization from a geo-parsed
  microblog stream}.
\newblock In {\it \bibinfo{booktitle}{Companion Publication of the 21st
  International Conference on Intelligent User Interfaces}\/} (pp.
  \bibinfo{pages}{58--61}).
\newblock \bibinfo{organization}{ACM}.
\bibitem[{Jurgens et~al.(2015)Jurgens, Finethy, McCorriston, Xu \&
  Ruths}]{jurgens2015geolocation}
\bibinfo{author}{Jurgens, D.}, \bibinfo{author}{Finethy, T.},
  \bibinfo{author}{McCorriston, J.}, \bibinfo{author}{Xu, Y.~T.}, \&
  \bibinfo{author}{Ruths, D.} (\bibinfo{year}{2015}).
\newblock \bibinfo{title}{Geolocation prediction in twitter using social
  networks: A critical analysis and review of current practice.}
\newblock {\it \bibinfo{journal}{ICWSM}\/},  {\it \bibinfo{volume}{15}\/},
  \bibinfo{pages}{188--197}.
\bibitem[{Kalchbrenner et~al.(2014)Kalchbrenner, Grefenstette \&
  Blunsom}]{kalchbrenner2014convolutional}
\bibinfo{author}{Kalchbrenner, N.}, \bibinfo{author}{Grefenstette, E.}, \&
  \bibinfo{author}{Blunsom, P.} (\bibinfo{year}{2014}).
\newblock \bibinfo{title}{A convolutional neural network for modelling
  sentences}.
\newblock {\it \bibinfo{journal}{arXiv preprint arXiv:1404.2188}\/}, .
\bibitem[{Karimi \& Yin(2012)}]{karimi2012microtext}
\bibinfo{author}{Karimi, S.}, \& \bibinfo{author}{Yin, J.}
  (\bibinfo{year}{2012}).
\newblock {\it \bibinfo{title}{Microtext annotation}\/}.
\newblock \bibinfo{type}{Technical Report} Technical Report EP13703, CSIRO.
\bibitem[{Kingma \& Ba(2014)}]{kingma2014adam}
\bibinfo{author}{Kingma, D.~P.}, \& \bibinfo{author}{Ba, J.}
  (\bibinfo{year}{2014}).
\newblock \bibinfo{title}{Adam: A method for stochastic optimization}.
\newblock {\it \bibinfo{journal}{arXiv preprint arXiv:1412.6980}\/}, .
\bibitem[{Kohavi et~al.(1995)}]{kohavi1995study}
\bibinfo{author}{Kohavi, R.} et~al. (\bibinfo{year}{1995}).
\newblock \bibinfo{title}{A study of cross-validation and bootstrap for
  accuracy estimation and model selection}.
\newblock In {\it \bibinfo{booktitle}{Ijcai}\/} (pp.
  \bibinfo{pages}{1137--1145}).
\newblock \bibinfo{organization}{Montreal, Canada} volume~\bibinfo{volume}{14}.
\bibitem[{Kumar et~al.(2017)Kumar, Singh \& Rana}]{kumar2017authenticity}
\bibinfo{author}{Kumar, A.}, \bibinfo{author}{Singh, J.~P.}, \&
  \bibinfo{author}{Rana, N.~P.} (\bibinfo{year}{2017}).
\newblock \bibinfo{title}{Authenticity of geo-location and place name in
  tweets}.
\newblock In {\it \bibinfo{booktitle}{23rd Americas Conference on Information
  Systems (AMCIS)}\/}.
\bibitem[{Lample et~al.(2016)Lample, Ballesteros, Subramanian, Kawakami \&
  Dyer}]{lample2016neural}
\bibinfo{author}{Lample, G.}, \bibinfo{author}{Ballesteros, M.},
  \bibinfo{author}{Subramanian, S.}, \bibinfo{author}{Kawakami, K.}, \&
  \bibinfo{author}{Dyer, C.} (\bibinfo{year}{2016}).
\newblock \bibinfo{title}{Neural architectures for named entity recognition}.
\newblock {\it \bibinfo{journal}{arXiv preprint arXiv:1603.01360}\/}, .
\bibitem[{Landwehr et~al.(2016)Landwehr, Wei, Kowalchuck \&
  Carley}]{landwehr2016using}
\bibinfo{author}{Landwehr, P.~M.}, \bibinfo{author}{Wei, W.},
  \bibinfo{author}{Kowalchuck, M.}, \& \bibinfo{author}{Carley, K.~M.}
  (\bibinfo{year}{2016}).
\newblock \bibinfo{title}{Using tweets to support disaster planning, warning
  and response}.
\newblock {\it \bibinfo{journal}{Safety science}\/},  {\it
  \bibinfo{volume}{90}\/}, \bibinfo{pages}{33--47}.
\bibitem[{Laylavi et~al.(2016)Laylavi, Rajabifard \&
  Kalantari}]{laylavi2016multi}
\bibinfo{author}{Laylavi, F.}, \bibinfo{author}{Rajabifard, A.}, \&
  \bibinfo{author}{Kalantari, M.} (\bibinfo{year}{2016}).
\newblock \bibinfo{title}{A multi-element approach to location inference of
  twitter: A case for emergency response}.
\newblock {\it \bibinfo{journal}{ISPRS International Journal of
  Geo-Information}\/},  {\it \bibinfo{volume}{5}\/}, \bibinfo{pages}{56}.
\bibitem[{Laylavi et~al.(2017)Laylavi, Rajabifard \&
  Kalantari}]{laylavi2017event}
\bibinfo{author}{Laylavi, F.}, \bibinfo{author}{Rajabifard, A.}, \&
  \bibinfo{author}{Kalantari, M.} (\bibinfo{year}{2017}).
\newblock \bibinfo{title}{Event relatedness assessment of twitter messages for
  emergency response}.
\newblock {\it \bibinfo{journal}{Information Processing \& Management}\/},
  {\it \bibinfo{volume}{53}\/}, \bibinfo{pages}{266--280}.
\bibitem[{Li \& Sun(2014)}]{li2014fine}
\bibinfo{author}{Li, C.}, \& \bibinfo{author}{Sun, A.} (\bibinfo{year}{2014}).
\newblock \bibinfo{title}{Fine-grained location extraction from tweets with
  temporal awareness}.
\newblock In {\it \bibinfo{booktitle}{Proceedings of the 37th international ACM
  SIGIR conference on Research \& development in information retrieval}\/} (pp.
  \bibinfo{pages}{43--52}).
\newblock \bibinfo{organization}{ACM}.
\bibitem[{Li et~al.(2012)Li, Weng, He, Yao, Datta, Sun \& Lee}]{li2012twiner}
\bibinfo{author}{Li, C.}, \bibinfo{author}{Weng, J.}, \bibinfo{author}{He, Q.},
  \bibinfo{author}{Yao, Y.}, \bibinfo{author}{Datta, A.}, \bibinfo{author}{Sun,
  A.}, \& \bibinfo{author}{Lee, B.-S.} (\bibinfo{year}{2012}).
\newblock \bibinfo{title}{Twiner: named entity recognition in targeted twitter
  stream}.
\newblock In {\it \bibinfo{booktitle}{Proceedings of the 35th international ACM
  SIGIR conference on Research and development in information retrieval}\/}
  (pp. \bibinfo{pages}{721--730}).
\newblock \bibinfo{organization}{ACM}.
\bibitem[{Limsopatham \& Collier(2016)}]{limsopatham2016bidirectional}
\bibinfo{author}{Limsopatham, N.}, \& \bibinfo{author}{Collier, N.~H.}
  (\bibinfo{year}{2016}).
\newblock \bibinfo{title}{Bidirectional lstm for named entity recognition in
  twitter messages}, .
\bibitem[{Lingad et~al.(2013)Lingad, Karimi \& Yin}]{lingad2013location}
\bibinfo{author}{Lingad, J.}, \bibinfo{author}{Karimi, S.}, \&
  \bibinfo{author}{Yin, J.} (\bibinfo{year}{2013}).
\newblock \bibinfo{title}{Location extraction from disaster-related
  microblogs}.
\newblock In {\it \bibinfo{booktitle}{Proceedings of the 22nd international
  conference on world wide web}\/} (pp. \bibinfo{pages}{1017--1020}).
\newblock \bibinfo{organization}{ACM}.
\bibitem[{Liu et~al.(2013)Liu, Wei, Zhang \& Zhou}]{liu2013named}
\bibinfo{author}{Liu, X.}, \bibinfo{author}{Wei, F.}, \bibinfo{author}{Zhang,
  S.}, \& \bibinfo{author}{Zhou, M.} (\bibinfo{year}{2013}).
\newblock \bibinfo{title}{Named entity recognition for tweets}.
\newblock {\it \bibinfo{journal}{ACM Transactions on Intelligent Systems and
  Technology (TIST)}\/},  {\it \bibinfo{volume}{4}\/}, \bibinfo{pages}{3}.
\bibitem[{Liu et~al.(2011)Liu, Zhang, Wei \& Zhou}]{liu2011recognizing}
\bibinfo{author}{Liu, X.}, \bibinfo{author}{Zhang, S.}, \bibinfo{author}{Wei,
  F.}, \& \bibinfo{author}{Zhou, M.} (\bibinfo{year}{2011}).
\newblock \bibinfo{title}{Recognizing named entities in tweets}.
\newblock In {\it \bibinfo{booktitle}{Proceedings of the 49th Annual Meeting of
  the Association for Computational Linguistics: Human Language
  Technologies-Volume 1}\/} (pp. \bibinfo{pages}{359--367}).
\newblock \bibinfo{organization}{Association for Computational Linguistics}.
\bibitem[{Lourentzou et~al.(2017)Lourentzou, Morales \&
  Zhai}]{lourentzou2017text}
\bibinfo{author}{Lourentzou, I.}, \bibinfo{author}{Morales, A.}, \&
  \bibinfo{author}{Zhai, C.} (\bibinfo{year}{2017}).
\newblock \bibinfo{title}{Text-based geolocation prediction of social media
  users with neural networks}.
\newblock In {\it \bibinfo{booktitle}{Big Data (Big Data), 2017 IEEE
  International Conference on}\/} (pp. \bibinfo{pages}{696--705}).
\newblock \bibinfo{organization}{IEEE}.
\bibitem[{Luna \& Pennock(2018)}]{luna2018social}
\bibinfo{author}{Luna, S.}, \& \bibinfo{author}{Pennock, M.~J.}
  (\bibinfo{year}{2018}).
\newblock \bibinfo{title}{Social media applications and emergency management: A
  literature review and research agenda}.
\newblock {\it \bibinfo{journal}{International Journal of Disaster Risk
  Reduction}\/}, .
\bibitem[{Malmasi \& Dras(2015)}]{malmasi2015location}
\bibinfo{author}{Malmasi, S.}, \& \bibinfo{author}{Dras, M.}
  (\bibinfo{year}{2015}).
\newblock \bibinfo{title}{Location mention detection in tweets and microblogs}.
\newblock In {\it \bibinfo{booktitle}{International Conference of the Pacific
  Association for Computational Linguistics}\/} (pp.
  \bibinfo{pages}{123--134}).
\newblock \bibinfo{organization}{Springer}.
\bibitem[{Mejri et~al.(2017)Mejri, Menoni, Matias \&
  Aminoltaheri}]{mejri2017crisis}
\bibinfo{author}{Mejri, O.}, \bibinfo{author}{Menoni, S.},
  \bibinfo{author}{Matias, K.}, \& \bibinfo{author}{Aminoltaheri, N.}
  (\bibinfo{year}{2017}).
\newblock \bibinfo{title}{Crisis information to support spatial planning in
  post disaster recovery}.
\newblock {\it \bibinfo{journal}{International Journal of Disaster Risk
  Reduction}\/},  {\it \bibinfo{volume}{22}\/}, \bibinfo{pages}{46--61}.
\bibitem[{Mendoza et~al.(2010)Mendoza, Poblete \&
  Castillo}]{mendoza2010twitter}
\bibinfo{author}{Mendoza, M.}, \bibinfo{author}{Poblete, B.}, \&
  \bibinfo{author}{Castillo, C.} (\bibinfo{year}{2010}).
\newblock \bibinfo{title}{Twitter under crisis: Can we trust what we rt?}
\newblock In {\it \bibinfo{booktitle}{Proceedings of the first workshop on
  social media analytics}\/} (pp. \bibinfo{pages}{71--79}).
\newblock \bibinfo{organization}{ACM}.
\bibitem[{Middleton et~al.(2018)Middleton, Kordopatis-Zilos, Papadopoulos \&
  Kompatsiaris}]{middleton2018location}
\bibinfo{author}{Middleton, S.}, \bibinfo{author}{Kordopatis-Zilos, G.},
  \bibinfo{author}{Papadopoulos, S.}, \& \bibinfo{author}{Kompatsiaris, Y.}
  (\bibinfo{year}{2018}).
\newblock \bibinfo{title}{Location extraction from social media:: geoparsing,
  location disambiguation and geotagging}.
\newblock {\it \bibinfo{journal}{ACM Transactions on Information Systems}\/}, .
\bibitem[{Middleton et~al.(2014)Middleton, Middleton \&
  Modafferi}]{middleton2014real}
\bibinfo{author}{Middleton, S.~E.}, \bibinfo{author}{Middleton, L.}, \&
  \bibinfo{author}{Modafferi, S.} (\bibinfo{year}{2014}).
\newblock \bibinfo{title}{Real-time crisis mapping of natural disasters using
  social media}.
\newblock {\it \bibinfo{journal}{IEEE Intelligent Systems}\/},  {\it
  \bibinfo{volume}{29}\/}, \bibinfo{pages}{9--17}.
\bibitem[{Mikolov et~al.(2013)Mikolov, Sutskever, Chen, Corrado \&
  Dean}]{mikolov2013distributed}
\bibinfo{author}{Mikolov, T.}, \bibinfo{author}{Sutskever, I.},
  \bibinfo{author}{Chen, K.}, \bibinfo{author}{Corrado, G.~S.}, \&
  \bibinfo{author}{Dean, J.} (\bibinfo{year}{2013}).
\newblock \bibinfo{title}{Distributed representations of words and phrases and
  their compositionality}.
\newblock In {\it \bibinfo{booktitle}{Advances in neural information processing
  systems}\/} (pp. \bibinfo{pages}{3111--3119}).
\bibitem[{Morstatter et~al.(2013)Morstatter, Pfeffer, Liu \&
  Carley}]{morstatter2013sample}
\bibinfo{author}{Morstatter, F.}, \bibinfo{author}{Pfeffer, J.},
  \bibinfo{author}{Liu, H.}, \& \bibinfo{author}{Carley, K.~M.}
  (\bibinfo{year}{2013}).
\newblock \bibinfo{title}{Is the sample good enough? comparing data from
  twitter's streaming api with twitter's firehose.}
\newblock In {\it \bibinfo{booktitle}{ICWSM}\/}.
\bibitem[{Nair \& Hinton(2010)}]{nair2010rectified}
\bibinfo{author}{Nair, V.}, \& \bibinfo{author}{Hinton, G.~E.}
  (\bibinfo{year}{2010}).
\newblock \bibinfo{title}{Rectified linear units improve restricted boltzmann
  machines}.
\newblock In {\it \bibinfo{booktitle}{Proceedings of the 27th international
  conference on machine learning (ICML-10)}\/} (pp. \bibinfo{pages}{807--814}).
\bibitem[{Nakaji \& Yanai(2012)}]{nakaji2012visualization}
\bibinfo{author}{Nakaji, Y.}, \& \bibinfo{author}{Yanai, K.}
  (\bibinfo{year}{2012}).
\newblock \bibinfo{title}{Visualization of real-world events with geotagged
  tweet photos}.
\newblock In {\it \bibinfo{booktitle}{Multimedia and Expo Workshops (ICMEW),
  2012 IEEE International Conference on}\/} (pp. \bibinfo{pages}{272--277}).
\newblock \bibinfo{organization}{IEEE}.
\bibitem[{Nam et~al.(2014)Nam, Kim, Menc{\'\i}a, Gurevych \&
  F{\"u}rnkranz}]{nam2014large}
\bibinfo{author}{Nam, J.}, \bibinfo{author}{Kim, J.},
  \bibinfo{author}{Menc{\'\i}a, E.~L.}, \bibinfo{author}{Gurevych, I.}, \&
  \bibinfo{author}{F{\"u}rnkranz, J.} (\bibinfo{year}{2014}).
\newblock \bibinfo{title}{Large-scale multi-label text
  classification—revisiting neural networks}.
\newblock In {\it \bibinfo{booktitle}{Joint european conference on machine
  learning and knowledge discovery in databases}\/} (pp.
  \bibinfo{pages}{437--452}).
\newblock \bibinfo{organization}{Springer}.
\bibitem[{Nguyen et~al.(2017{\natexlab{a}})Nguyen, Al-Mannai, Joty, Sajjad,
  Imran \& Mitra}]{nguyen2017robust}
\bibinfo{author}{Nguyen, D.~T.}, \bibinfo{author}{Al-Mannai, K.},
  \bibinfo{author}{Joty, S.~R.}, \bibinfo{author}{Sajjad, H.},
  \bibinfo{author}{Imran, M.}, \& \bibinfo{author}{Mitra, P.}
  (\bibinfo{year}{2017}{\natexlab{a}}).
\newblock \bibinfo{title}{Robust classification of crisis-related data on
  social networks using convolutional neural networks.}
\newblock In {\it \bibinfo{booktitle}{ICWSM}\/} (pp.
  \bibinfo{pages}{632--635}).
\bibitem[{Nguyen et~al.(2017{\natexlab{b}})Nguyen, Yang, Kim \&
  Oh}]{nguyen2017real}
\bibinfo{author}{Nguyen, V.~Q.}, \bibinfo{author}{Yang, H.-J.},
  \bibinfo{author}{Kim, K.}, \& \bibinfo{author}{Oh, A.-R.}
  (\bibinfo{year}{2017}{\natexlab{b}}).
\newblock \bibinfo{title}{Real-time earthquake detection using convolutional
  neural network and social data}.
\newblock In {\it \bibinfo{booktitle}{Multimedia Big Data (BigMM), 2017 IEEE
  Third International Conference on}\/} (pp. \bibinfo{pages}{154--157}).
\newblock \bibinfo{organization}{IEEE}.
\bibitem[{Olteanu et~al.(2014)Olteanu, Castillo, Diaz \&
  Vieweg}]{olteanu2014crisislex}
\bibinfo{author}{Olteanu, A.}, \bibinfo{author}{Castillo, C.},
  \bibinfo{author}{Diaz, F.}, \& \bibinfo{author}{Vieweg, S.}
  (\bibinfo{year}{2014}).
\newblock \bibinfo{title}{Crisislex: A lexicon for collecting and filtering
  microblogged communications in crises.}
\newblock In {\it \bibinfo{booktitle}{ICWSM}\/}.
\bibitem[{Olteanu et~al.(2015)Olteanu, Vieweg \& Castillo}]{olteanu2015expect}
\bibinfo{author}{Olteanu, A.}, \bibinfo{author}{Vieweg, S.}, \&
  \bibinfo{author}{Castillo, C.} (\bibinfo{year}{2015}).
\newblock \bibinfo{title}{What to expect when the unexpected happens: Social
  media communications across crises}.
\newblock In {\it \bibinfo{booktitle}{Proceedings of the 18th ACM Conference on
  Computer Supported Cooperative Work \& Social Computing}\/} (pp.
  \bibinfo{pages}{994--1009}).
\newblock \bibinfo{organization}{ACM}.
\bibitem[{Ozdikis et~al.(2017)Ozdikis, O{\u{g}}uzt{\"u}z{\"u}n \&
  Karagoz}]{ozdikis2017survey}
\bibinfo{author}{Ozdikis, O.}, \bibinfo{author}{O{\u{g}}uzt{\"u}z{\"u}n, H.},
  \& \bibinfo{author}{Karagoz, P.} (\bibinfo{year}{2017}).
\newblock \bibinfo{title}{A survey on location estimation techniques for events
  detected in twitter}.
\newblock {\it \bibinfo{journal}{Knowledge and Information Systems}\/},  {\it
  \bibinfo{volume}{52}\/}, \bibinfo{pages}{291--339}.
\bibitem[{Panteras et~al.(2015)Panteras, Wise, Lu, Croitoru, Crooks \&
  Stefanidis}]{panteras2015triangulating}
\bibinfo{author}{Panteras, G.}, \bibinfo{author}{Wise, S.},
  \bibinfo{author}{Lu, X.}, \bibinfo{author}{Croitoru, A.},
  \bibinfo{author}{Crooks, A.}, \& \bibinfo{author}{Stefanidis, A.}
  (\bibinfo{year}{2015}).
\newblock \bibinfo{title}{Triangulating social multimedia content for event
  localization using flickr and twitter}.
\newblock {\it \bibinfo{journal}{Transactions in GIS}\/},  {\it
  \bibinfo{volume}{19}\/}, \bibinfo{pages}{694--715}.
\bibitem[{Pennington et~al.(2014)Pennington, Socher \&
  Manning}]{pennington2014glove}
\bibinfo{author}{Pennington, J.}, \bibinfo{author}{Socher, R.}, \&
  \bibinfo{author}{Manning, C.} (\bibinfo{year}{2014}).
\newblock \bibinfo{title}{Glove: Global vectors for word representation}.
\newblock In {\it \bibinfo{booktitle}{Proceedings of the 2014 conference on
  empirical methods in natural language processing (EMNLP)}\/} (pp.
  \bibinfo{pages}{1532--1543}).
\bibitem[{Perol et~al.(2018)Perol, Gharbi \& Denolle}]{perol2018convolutional}
\bibinfo{author}{Perol, T.}, \bibinfo{author}{Gharbi, M.}, \&
  \bibinfo{author}{Denolle, M.} (\bibinfo{year}{2018}).
\newblock \bibinfo{title}{Convolutional neural network for earthquake detection
  and location}.
\newblock {\it \bibinfo{journal}{Science Advances}\/},  {\it
  \bibinfo{volume}{4}\/}, \bibinfo{pages}{e1700578}.
\bibitem[{Qian et~al.(2017)Qian, Tang, Yang, Huang, Wei \&
  Carley}]{qian2017probabilistic}
\bibinfo{author}{Qian, Y.}, \bibinfo{author}{Tang, J.}, \bibinfo{author}{Yang,
  Z.}, \bibinfo{author}{Huang, B.}, \bibinfo{author}{Wei, W.}, \&
  \bibinfo{author}{Carley, K.~M.} (\bibinfo{year}{2017}).
\newblock \bibinfo{title}{A probabilistic framework for location inference from
  social media}.
\newblock {\it \bibinfo{journal}{arXiv preprint arXiv:1702.07281}\/}, .
\bibitem[{Ritter et~al.(2011)Ritter, Clark, Etzioni et~al.}]{ritter2011named}
\bibinfo{author}{Ritter, A.}, \bibinfo{author}{Clark, S.},
  \bibinfo{author}{Etzioni, O.} et~al. (\bibinfo{year}{2011}).
\newblock \bibinfo{title}{Named entity recognition in tweets: an experimental
  study}.
\newblock In {\it \bibinfo{booktitle}{Proceedings of the conference on
  empirical methods in natural language processing}\/} (pp.
  \bibinfo{pages}{1524--1534}).
\newblock \bibinfo{organization}{Association for Computational Linguistics}.
\bibitem[{Rossi et~al.(2018)Rossi, Acerbo, Ylinen, Juga, Nurmi, Bosca,
  Tarasconi, Cristoforetti \& Alikadic}]{rossi2018early}
\bibinfo{author}{Rossi, C.}, \bibinfo{author}{Acerbo, F.},
  \bibinfo{author}{Ylinen, K.}, \bibinfo{author}{Juga, I.},
  \bibinfo{author}{Nurmi, P.}, \bibinfo{author}{Bosca, A.},
  \bibinfo{author}{Tarasconi, F.}, \bibinfo{author}{Cristoforetti, M.}, \&
  \bibinfo{author}{Alikadic, A.} (\bibinfo{year}{2018}).
\newblock \bibinfo{title}{Early detection and information extraction for
  weather-induced floods using social media streams}.
\newblock {\it \bibinfo{journal}{International Journal of Disaster Risk
  Reduction}\/}, .
\bibitem[{Sakaki et~al.(2013)Sakaki, Okazaki \& Matsuo}]{sakaki2013tweet}
\bibinfo{author}{Sakaki, T.}, \bibinfo{author}{Okazaki, M.}, \&
  \bibinfo{author}{Matsuo, Y.} (\bibinfo{year}{2013}).
\newblock \bibinfo{title}{Tweet analysis for real-time event detection and
  earthquake reporting system development}.
\newblock {\it \bibinfo{journal}{IEEE Transactions on Knowledge and Data
  Engineering}\/},  {\it \bibinfo{volume}{25}\/}, \bibinfo{pages}{919--931}.
\bibitem[{Sankaranarayanan et~al.(2009)Sankaranarayanan, Samet, Teitler,
  Lieberman \& Sperling}]{sankaranarayanan2009twitterstand}
\bibinfo{author}{Sankaranarayanan, J.}, \bibinfo{author}{Samet, H.},
  \bibinfo{author}{Teitler, B.~E.}, \bibinfo{author}{Lieberman, M.~D.}, \&
  \bibinfo{author}{Sperling, J.} (\bibinfo{year}{2009}).
\newblock \bibinfo{title}{Twitterstand: news in tweets}.
\newblock In {\it \bibinfo{booktitle}{Proceedings of the 17th acm sigspatial
  international conference on advances in geographic information systems}\/}
  (pp. \bibinfo{pages}{42--51}).
\newblock \bibinfo{organization}{ACM}.
\bibitem[{Shibuya(2017)}]{shibuya2017mining}
\bibinfo{author}{Shibuya, Y.} (\bibinfo{year}{2017}).
\newblock \bibinfo{title}{Mining social media for disaster management:
  Leveraging social media data for community recovery}.
\newblock In {\it \bibinfo{booktitle}{Big Data (Big Data), 2017 IEEE
  International Conference on}\/} (pp. \bibinfo{pages}{3111--3118}).
\newblock \bibinfo{organization}{IEEE}.
\bibitem[{Sikdar \& Gamb{\"a}ck(2016)}]{sikdar2016feature}
\bibinfo{author}{Sikdar, U.~K.}, \& \bibinfo{author}{Gamb{\"a}ck, B.}
  (\bibinfo{year}{2016}).
\newblock \bibinfo{title}{Feature-rich twitter named entity recognition and
  classification}.
\newblock In {\it \bibinfo{booktitle}{Proceedings of the 2nd Workshop on Noisy
  User-generated Text (WNUT)}\/} (pp. \bibinfo{pages}{164--170}).
\bibitem[{Singh et~al.(2017)Singh, Dwivedi, Rana, Kumar \&
  Kapoor}]{singh2017event}
\bibinfo{author}{Singh, J.~P.}, \bibinfo{author}{Dwivedi, Y.~K.},
  \bibinfo{author}{Rana, N.~P.}, \bibinfo{author}{Kumar, A.}, \&
  \bibinfo{author}{Kapoor, K.~K.} (\bibinfo{year}{2017}).
\newblock \bibinfo{title}{Event classification and location prediction from
  tweets during disasters}.
\newblock {\it \bibinfo{journal}{Annals of Operations Research}\/},  (pp.
  \bibinfo{pages}{1--21}).
\bibitem[{Socher et~al.(2013)Socher, Perelygin, Wu, Chuang, Manning, Ng \&
  Potts}]{socher2013recursive}
\bibinfo{author}{Socher, R.}, \bibinfo{author}{Perelygin, A.},
  \bibinfo{author}{Wu, J.}, \bibinfo{author}{Chuang, J.},
  \bibinfo{author}{Manning, C.~D.}, \bibinfo{author}{Ng, A.}, \&
  \bibinfo{author}{Potts, C.} (\bibinfo{year}{2013}).
\newblock \bibinfo{title}{Recursive deep models for semantic compositionality
  over a sentiment treebank}.
\newblock In {\it \bibinfo{booktitle}{Proceedings of the 2013 conference on
  empirical methods in natural language processing}\/} (pp.
  \bibinfo{pages}{1631--1642}).
\bibitem[{Srivastava et~al.(2014)Srivastava, Hinton, Krizhevsky, Sutskever \&
  Salakhutdinov}]{srivastava2014dropout}
\bibinfo{author}{Srivastava, N.}, \bibinfo{author}{Hinton, G.},
  \bibinfo{author}{Krizhevsky, A.}, \bibinfo{author}{Sutskever, I.}, \&
  \bibinfo{author}{Salakhutdinov, R.} (\bibinfo{year}{2014}).
\newblock \bibinfo{title}{Dropout: A simple way to prevent neural networks from
  overfitting}.
\newblock {\it \bibinfo{journal}{The Journal of Machine Learning Research}\/},
  {\it \bibinfo{volume}{15}\/}, \bibinfo{pages}{1929--1958}.
\bibitem[{Temnikova et~al.(2015)Temnikova, Vieweg \&
  Castillo}]{temnikova2015case}
\bibinfo{author}{Temnikova, I.}, \bibinfo{author}{Vieweg, S.}, \&
  \bibinfo{author}{Castillo, C.} (\bibinfo{year}{2015}).
\newblock \bibinfo{title}{The case for readability of crisis communications in
  social media}.
\newblock In {\it \bibinfo{booktitle}{Proceedings of the 24th international
  conference on world wide web}\/} (pp. \bibinfo{pages}{1245--1250}).
\newblock \bibinfo{organization}{ACM}.
\bibitem[{Unankard et~al.(2015)Unankard, Li \& Sharaf}]{unankard2015emerging}
\bibinfo{author}{Unankard, S.}, \bibinfo{author}{Li, X.}, \&
  \bibinfo{author}{Sharaf, M.~A.} (\bibinfo{year}{2015}).
\newblock \bibinfo{title}{Emerging event detection in social networks with
  location sensitivity}.
\newblock {\it \bibinfo{journal}{World Wide Web}\/},  {\it
  \bibinfo{volume}{18}\/}, \bibinfo{pages}{1393--1417}.
\bibitem[{Vieweg et~al.(2010)Vieweg, Hughes, Starbird \&
  Palen}]{vieweg2010microblogging}
\bibinfo{author}{Vieweg, S.}, \bibinfo{author}{Hughes, A.~L.},
  \bibinfo{author}{Starbird, K.}, \& \bibinfo{author}{Palen, L.}
  (\bibinfo{year}{2010}).
\newblock \bibinfo{title}{Microblogging during two natural hazards events: what
  twitter may contribute to situational awareness}.
\newblock In {\it \bibinfo{booktitle}{Proceedings of the SIGCHI conference on
  human factors in computing systems}\/} (pp. \bibinfo{pages}{1079--1088}).
\newblock \bibinfo{organization}{ACM}.
\bibitem[{Xu et~al.(2015)Xu, Chow, Yiu, Li \& Poon}]{xu2015mobifeed}
\bibinfo{author}{Xu, W.}, \bibinfo{author}{Chow, C.-Y.}, \bibinfo{author}{Yiu,
  M.~L.}, \bibinfo{author}{Li, Q.}, \& \bibinfo{author}{Poon, C.~K.}
  (\bibinfo{year}{2015}).
\newblock \bibinfo{title}{Mobifeed: A location-aware news feed framework for
  moving users}.
\newblock {\it \bibinfo{journal}{GeoInformatica}\/},  {\it
  \bibinfo{volume}{19}\/}, \bibinfo{pages}{633--669}.
\bibitem[{Yang et~al.(2017)Yang, Nguyen, Stuve, Cao \& Jin}]{yang2017harvey}
\bibinfo{author}{Yang, Z.}, \bibinfo{author}{Nguyen, L.~H.},
  \bibinfo{author}{Stuve, J.}, \bibinfo{author}{Cao, G.}, \&
  \bibinfo{author}{Jin, F.} (\bibinfo{year}{2017}).
\newblock \bibinfo{title}{Harvey flooding rescue in social media}.
\newblock In {\it \bibinfo{booktitle}{Big Data (Big Data), 2017 IEEE
  International Conference on}\/} (pp. \bibinfo{pages}{2177--2185}).
\newblock \bibinfo{organization}{IEEE}.
\bibitem[{Yuan \& Liu(2018)}]{yuan2018feasibility}
\bibinfo{author}{Yuan, F.}, \& \bibinfo{author}{Liu, R.}
  (\bibinfo{year}{2018}).
\newblock \bibinfo{title}{Feasibility study of using crowdsourcing to identify
  critical affected areas for rapid damage assessment: Hurricane matthew case
  study}.
\newblock {\it \bibinfo{journal}{International journal of disaster risk
  reduction}\/},  {\it \bibinfo{volume}{28}\/}, \bibinfo{pages}{758--767}.
\bibitem[{Yuan et~al.(2013)Yuan, Cong, Ma, Sun \& Thalmann}]{yuan2013and}
\bibinfo{author}{Yuan, Q.}, \bibinfo{author}{Cong, G.}, \bibinfo{author}{Ma,
  Z.}, \bibinfo{author}{Sun, A.}, \& \bibinfo{author}{Thalmann, N.~M.}
  (\bibinfo{year}{2013}).
\newblock \bibinfo{title}{Who, where, when and what: discover spatio-temporal
  topics for twitter users}.
\newblock In {\it \bibinfo{booktitle}{Proceedings of the 19th ACM SIGKDD
  international conference on Knowledge discovery and data mining}\/} (pp.
  \bibinfo{pages}{605--613}).
\newblock \bibinfo{organization}{ACM}.
\bibitem[{Zhang \& Gelernter(2014)}]{zhang2014geocoding}
\bibinfo{author}{Zhang, W.}, \& \bibinfo{author}{Gelernter, J.}
  (\bibinfo{year}{2014}).
\newblock \bibinfo{title}{Geocoding location expressions in twitter messages: A
  preference learning method}.
\newblock {\it \bibinfo{journal}{Journal of Spatial Information Science}\/},
  {\it \bibinfo{volume}{2014}\/}, \bibinfo{pages}{37--70}.
\bibitem[{Zheng et~al.(2018)Zheng, Han \& Sun}]{zheng2018survey}
\bibinfo{author}{Zheng, X.}, \bibinfo{author}{Han, J.}, \&
  \bibinfo{author}{Sun, A.} (\bibinfo{year}{2018}).
\newblock \bibinfo{title}{A survey of location prediction on twitter}.
\newblock {\it \bibinfo{journal}{IEEE Transactions on Knowledge and Data
  Engineering}\/}, .
\bibitem[{Zhou et~al.(2017)Zhou, Wu, Xu \& Fujita}]{zhou2017emergency}
\bibinfo{author}{Zhou, L.}, \bibinfo{author}{Wu, X.}, \bibinfo{author}{Xu, Z.},
  \& \bibinfo{author}{Fujita, H.} (\bibinfo{year}{2017}).
\newblock \bibinfo{title}{Emergency decision making for natural disasters: An
  overview}.
\newblock {\it \bibinfo{journal}{International Journal of Disaster Risk
  Reduction}\/}, .

\end{thebibliography}
\end{document}